\documentclass{article}

    \PassOptionsToPackage{numbers, compress}{natbib}
\usepackage[preprint]{neurips_2026}

\usepackage{tabularx}
\usepackage[utf8]{inputenc} 
\usepackage[T1]{fontenc}    
\usepackage{hyperref}       
\usepackage{url}            
\usepackage{booktabs}       
\usepackage{amsfonts}       
\usepackage{nicefrac}       
\usepackage{microtype}      

\usepackage{amsmath}
\usepackage{bbm}
\usepackage{longtable}
\usepackage{booktabs}
\usepackage{float}
\usepackage{multirow}
\usepackage{graphicx}
\usepackage{cleveref}
\usepackage{placeins}
\usepackage{amssymb}
\usepackage{pifont}
\usepackage{wrapfig}
\usepackage{subcaption}
\usepackage[table]{xcolor}

\newcommand{\cmarkg}{\textcolor{green!60!black}{\ding{51}}}
\newcommand{\xmarkr}{\textcolor{red!80!black}{\ding{55}}}

\title{Beyond Classification: Dynamic Adapter Routing\\for Continual Multimodal Retrieval}

\author{%
  Alicja Dobrzeniecka\thanks{Email: \texttt{alicja.dobrzeniecka@nask.pl}} \\
  NASK National Research Institute \\
  \And
  Filip Szatkowski \\
  IDEAS Research Institute \\
  Warsaw University of Technology \\
  \AND
  Sebastian Cygert \\
  NASK National Research Institute \\
  \And
  Szymon Łukasik \\
  NASK National Research Institute \\
  \And
  Bartłomiej Twardowski \\
  Universitat Autonoma de Barcelona \\
  IDEAS Research Institute\\
}

\begin{document}

\maketitle

\begin{abstract}
While retrieval is a core function of vision-language models, continually updating these models for retrieval tasks remains critically underexplored. 
Existing work often approaches continual retrieval through the lens of class-incremental learning~(CIL), evaluating both standard CIL methods and retrieval-oriented adaptations in settings that may not fully capture the retrieval-specific dynamics. 
To address this, we introduce a new, principled evaluation framework for continual multimodal retrieval~(CMR) spanning diverse visual domains, and systematically evaluate common approaches within this setting. Our empirical analysis shows that standard CIL methods fail to yield meaningful gains in our more challenging scenario. 
Therefore, we propose Dynamic Adapter Routing~(DAR), a novel approach based on adapters selected through prototype-based routing and combined via model merging.DAR achieves superior performance over the previous baselines and demonstrates strong generalization under out-of-distribution evaluation. 
Our results highlights the unique challenges of CMR and encourages further research in this direction.
\end{abstract}

\section{Introduction}
\begin{figure}[t]
    \centering
    \hspace{0.5cm}
    \includegraphics[width=0.2\textwidth]{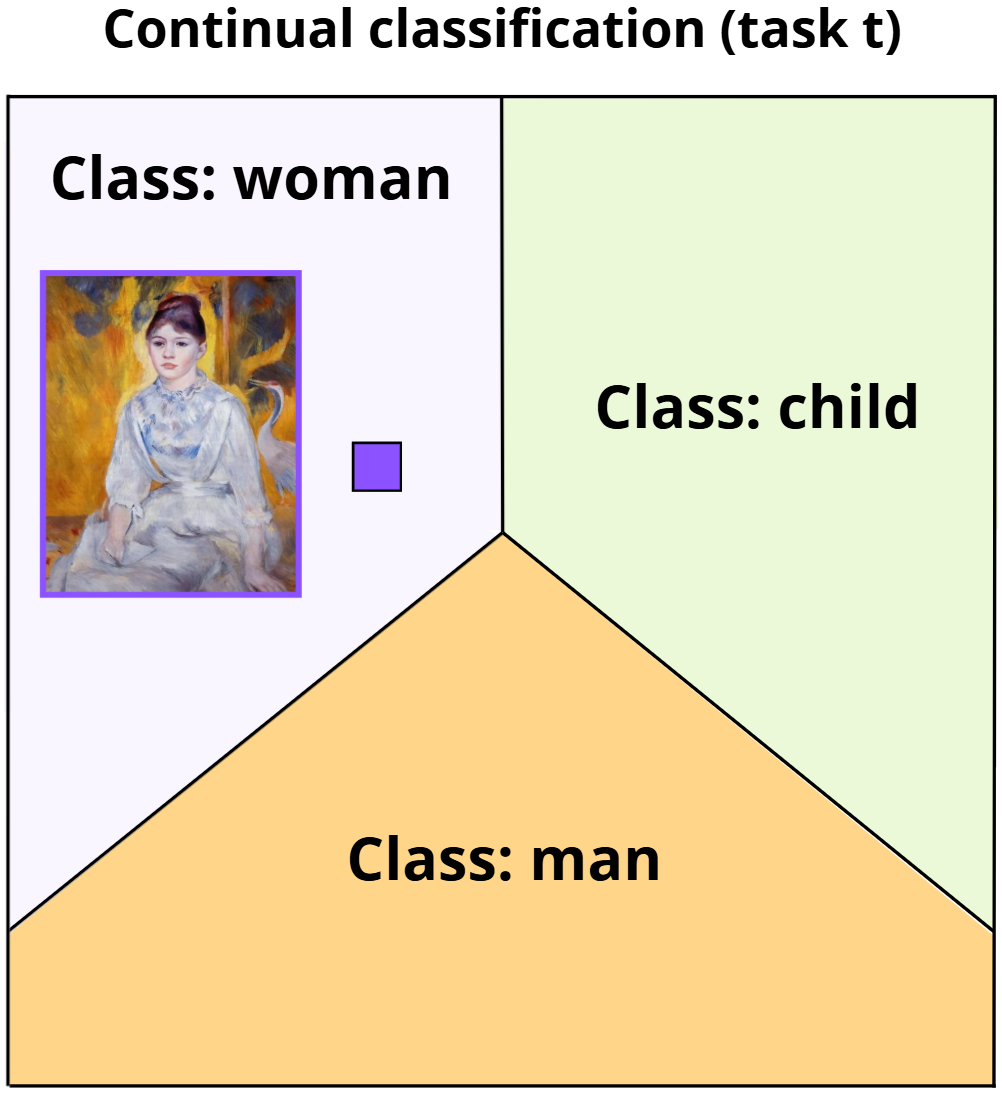} 
    \hfill
    \includegraphics[width=0.2\textwidth]{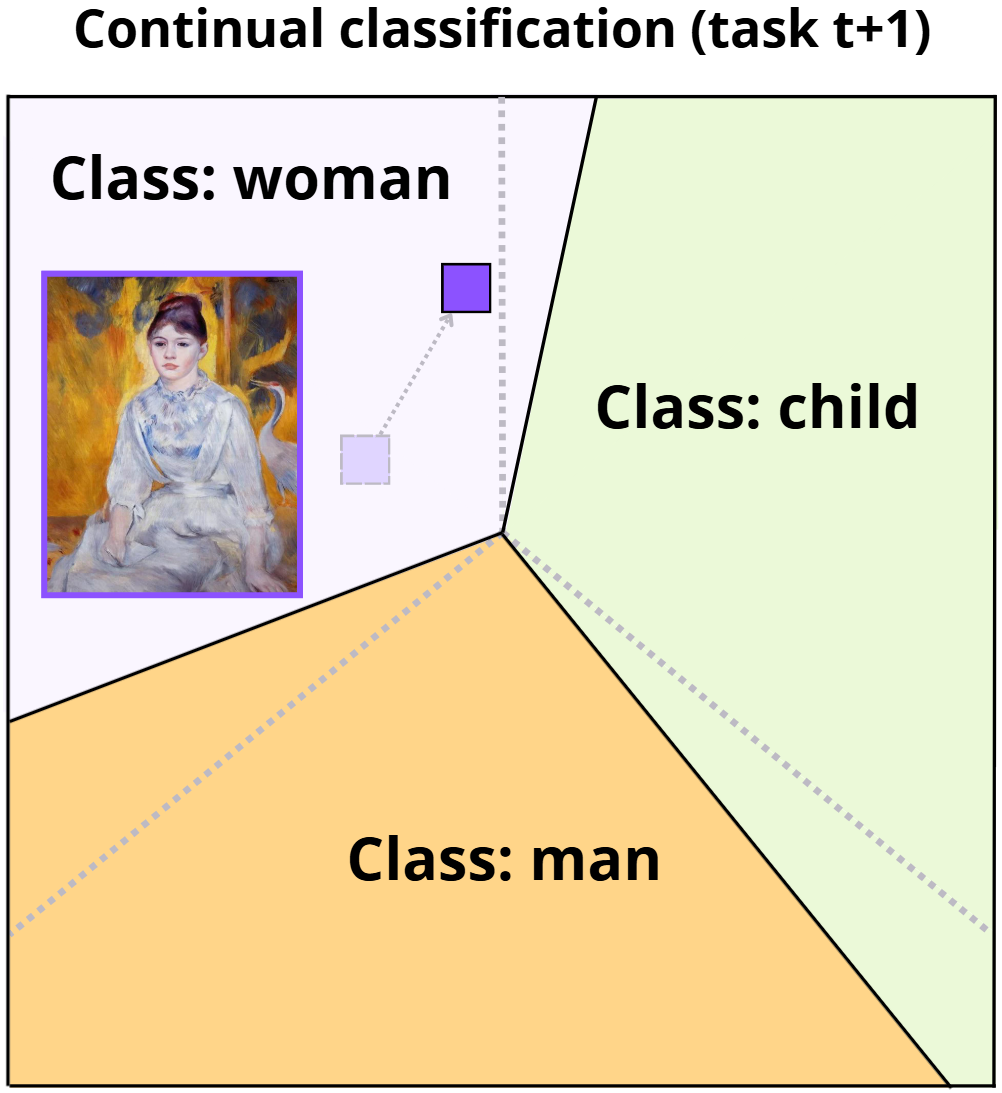} 
    \hfill
    \includegraphics[width=0.2\textwidth]{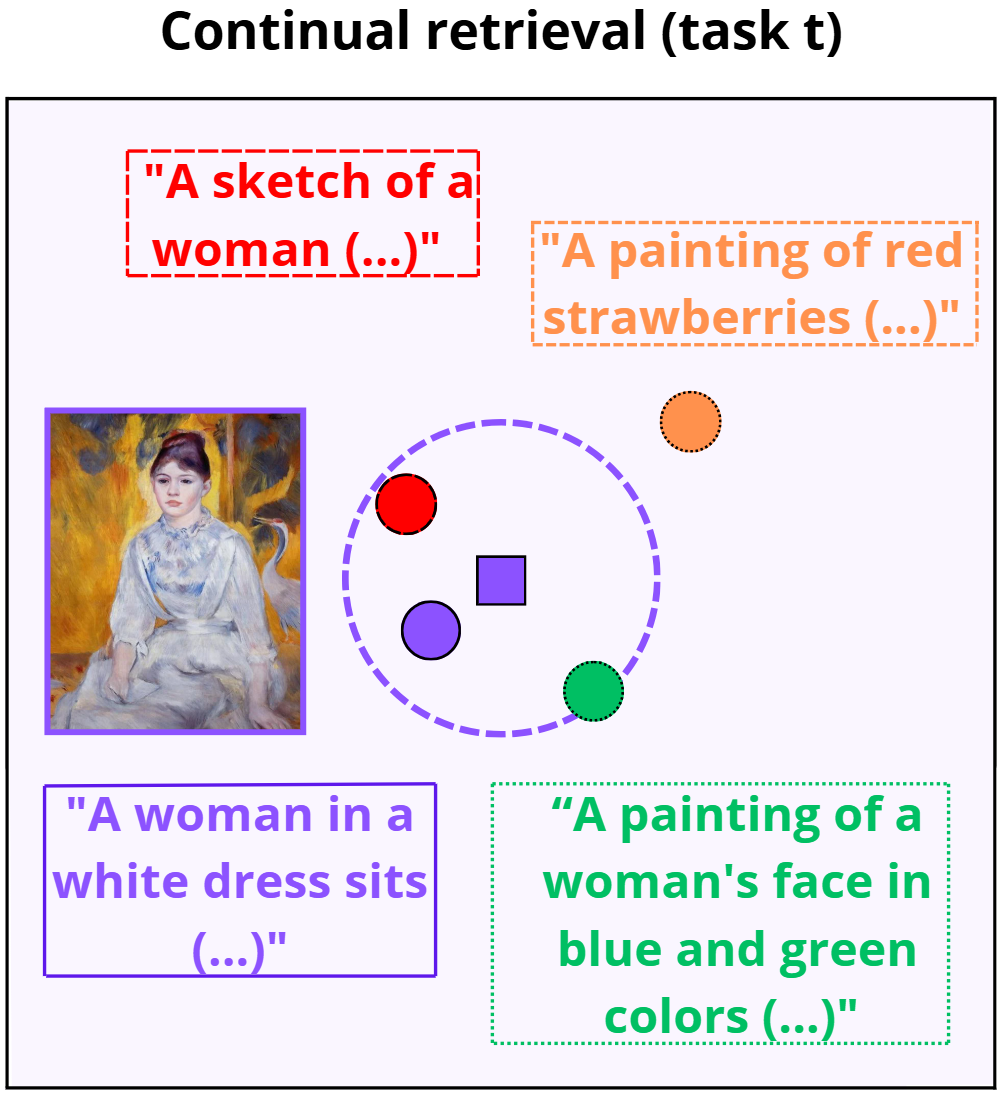} 
    \hfill
    \includegraphics[width=0.2\textwidth]{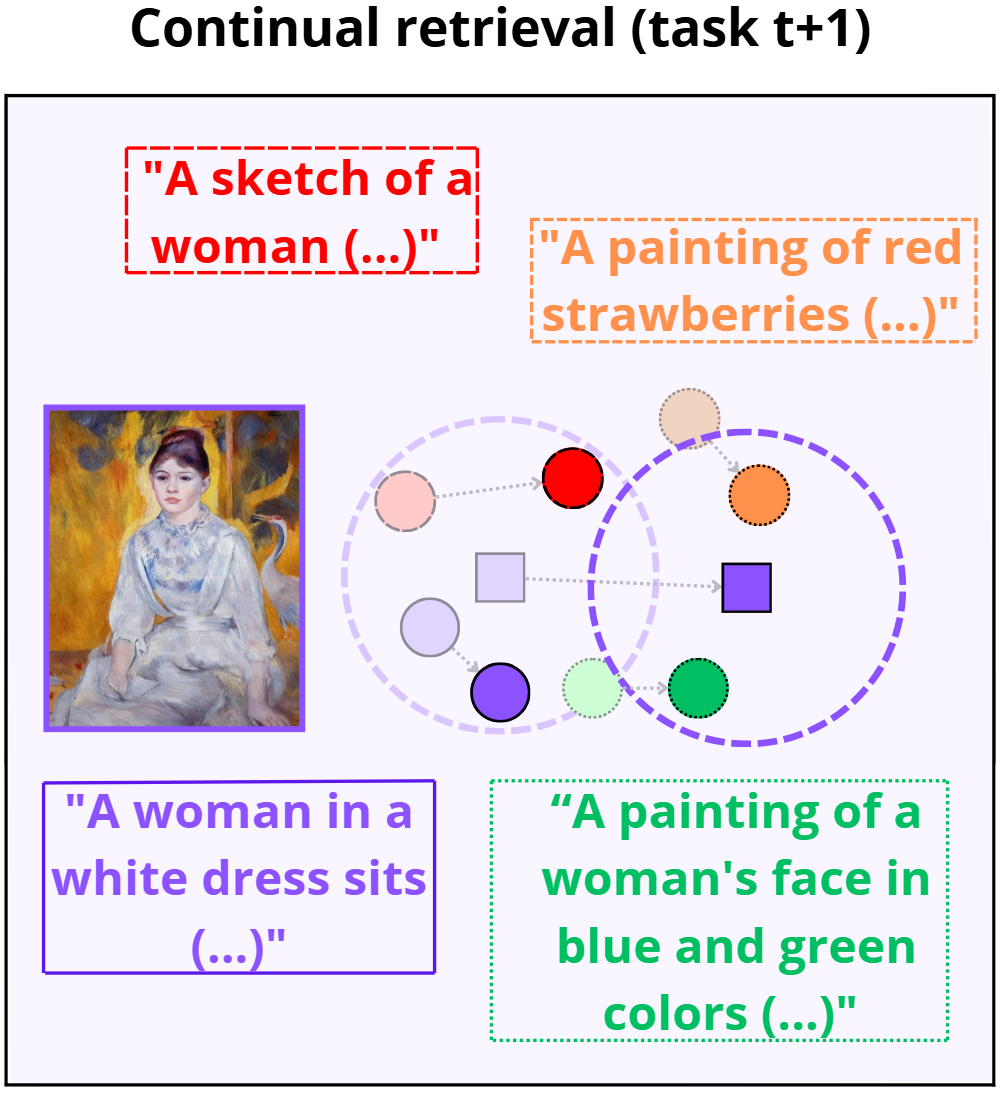}
    \hspace{0.5cm}
    \caption{Conceptual difference between classification and retrieval in CL scenario. In classification, small perturbations may not affect the result as long as the sample remains within the class boundaries. 
    In retrieval, even small perturbations can alter nearest neighbours and substantially affect retrieval rankings. 
    We argue that continual retrieval requires dedicated evaluation protocols and methods, and introduce a new retrieval-focused benchmark and a novel, state-of-the-art method.\vspace{-0.5cm}
    }
    \label{fig:teaser}
\end{figure}

Retrieval is a core functionality of vision-language models~(VLMs) such as CLIP~\citep{radford2021clip}, serving as the primary interface for deploying these models in real-world search, recommendation, and indexing systems. 
Despite its immense practical importance, the problem of continually updating multimodal models for retrieval has received relatively limited attention. Only a small number of works explicitly address continual cross-modal retrieval~\citep{c_clip2025, cui2024knowledge_rectification, ni2023off_diagonal, li2025coleclip}, while the majority of research on continual learning~(CL) for VLMs focuses on classification-oriented settings~\citep{Wang2021LearningTP, jha2024clapclip, yu2024boosting, huang2024rapf, lu2025continual, zheng2023preventing_zero_shot}, particularly class-incremental learning~(CIL). 
As a result, existing approaches fail to capture the unique dynamics and challenges inherent to multimodal retrieval, which requires maintaining a globally consistent embedding space
~\citep{kai2021cross_modal_retrieval} instead of merely learning discrete decision boundaries.  
This creates unique failure modes: representation drift can ruin global rankings even if a model learns a new task perfectly, and catastrophic forgetting distorts relative similarities rather than causing simple misclassifications~\citep{cui2024knowledge_rectification, ni2023off_diagonal} as we present conceptually in \Cref{fig:teaser}. 

Furthermore, existing works on continual retrieval suffer from limited scale and domain diversity in their evaluations, failing to present sufficiently retrieval-specific scenarios. 
This in turn produces overly optimistic assessments, ultimately obscuring the true utility of CL approaches.
To address this, we introduce a new evaluation framework for continual multimodal retrieval. 
Our framework includes sequences of heterogeneous, non-overlapping datasets that span various visual domains.
This design ensures a sufficiently challenging evaluation, enabling more principled comparison. 
Additionally, we assess the model performance on out-of-distribution~(OOD) data, and demonstrate that improving both in- and out-of-distribution results remains challenging for the commonly used continual methods.
We use our framework to conduct a systematic analysis of knowledge transfer, interference and robustness under realistic distribution shifts across common CIL methods and retrieval-oriented approaches from literature. Surprisingly, we find that commonly used CL strategies fail to deliver consistent improvements in our more challenging setup.

Motivated by these findings, we introduce Dynamic Adapter Routing (DAR), a novel method for continual multimodal retrieval. Unlike existing routing and mixture-of-experts approaches designed for classification or task-aware settings, continual retrieval requires preserving a globally consistent embedding space under ambiguous cross-domain queries. DAR addresses this challenge through retrieval-aware prototype routing and uncertainty-triggered adapter merging, enabling adaptive knowledge transfer while limiting representation drift.

We perform exhaustive ablations of our method and demonstrate that it achieves superior results over the prior CL baselines across heterogeneous retrieval benchmarks, while retaining strong flexibility under OOD evaluation.
Our main contributions include:
\vspace{-0.2cm}

\begin{itemize}
    \item \textbf{New, more challenging benchmark for continual multimodal retrieval.} We present a CL framework focused on retrieval, comprising heterogeneous, non-overlapping domains, alongside in-distribution and out-of-distribution evaluation protocols.\vspace{-0.1cm}
    \item \textbf{Systematic evaluation of CL methods for continual retrieval.} Within our framework, we evaluate common CIL and retrieval-oriented approaches, demonstrating that many existing methods offer limited improvements in more demanding settings. We further analyse the correlation between embedding-space alignment and structure and retrieval performance.\vspace{-0.1cm}
    \item \textbf{Dynamic Adapter Routing (DAR).} We introduce a continual retrieval method that uses prototype-guided routing and adaptive merging to mitigate representation drift and cross-task interference. DAR significantly outperforms prior approaches and stands out as the only method that delivers consistent, meaningful improvements.\vspace{-0.2cm}
\end{itemize}

\section{Related Work}

\paragraph{CL for VLMs.}

Recent research in CL for VLMs has primarily focused on maintaining classification accuracy and zero-shot capabilities under distribution shifts. 
Representative approaches include ZSCL~\citep{zheng2023preventing_zero_shot}, which employs distillation from a frozen pre-trained backbone to mitigate forgetting, and various prompt-based methods~\citep{smith2023construct_vl, yu2024boosting} that utilize learnable tokens to bypass the risks of full fine-tuning. 
While these methods offer valuable insights into stabilizing classification, they are fundamentally optimized for discriminative objectives, which may 
not transfer well to retrieval settings, where the preservation of fine-grained, cross-modal alignment is paramount.
A specialized subset of literature addresses the unique challenges of continual cross-modal retrieval. 
\citet{kai2021cross_modal_retrieval} establish that maintaining a coherent shared embedding space across tasks is a distinct requirement that separates retrieval from standard CIL. 
Subsequent methods like DKR~\citep{cui2024knowledge_rectification}, Mod-X~\citep{ni2023off_diagonal}, and C-CLIP \citep{c_clip2025} have attempted to mitigate representation drift through architectural adaptations, dynamic knowledge updates, or parameter-efficient fine-tuning (PEFT). 

\vspace{-0.2cm}

\paragraph{Continual Retrieval Benchmarks.}
Historically, CL for VLMs has primarily relied on CIL benchmarks (e.g., CIFAR~\cite{krizhevsky2009learning} or ImageNet~\cite{imagenet} splits), which are inherently classification-oriented and therefore inadequately capture retrieval-specific performance. 
Some works propose the frameworks for retrieval evaluation, but they are fundamentally insufficient.
While TiC-CLIP~\citep{garg2024tic_clip} evaluates retrieval on temporal data streams, it focuses on large-scale continual pretraining settings that may be less practical for lightweight and reproducible downstream evaluation.
Similarly, FOMO-in-FLUX~\citep{udandaro2024fomo_in_flux} primarily targets pretraining dynamics rather than practical downstream adaptation. 
Although C-CLIP~\citep{c_clip2025} recently proposed a framework for retrieval evaluation, its reliance on private datasets and a closed-source implementation hinders reproducibility and broader community adoption.
Overall, existing evaluations often rely on datasets with limited domain diversity, potentially obscuring retrieval-specific failure modes and leading to overly optimistic performance estimates. 
This motivates the need for a more principled evaluation framework, which we introduce and discuss in detail in \Cref{sec:framework}.
\vspace{-0.2cm}

\paragraph{Parameter-Efficient Adaptation and Model Routing.}
PEFT methods, such as LoRA~\citep{hu2022lowrank} and adapter-based modules~\cite{pfeiffer-etal-2020-adapterhub}, have emerged as the standard for adapting large-scale models with minimal overhead. 
To scale these methods to multiple tasks in CL, recent work has explored MoE or adapter-based routing, which maintain task-specific modules selected dynamically at inference~\citep{yu2024boosting}. 
More recently, model merging techniques have been proposed to consolidate multiple task-specific adaptations into a unified model without increasing the inference-time parameter count, and proved effective for CL~\citep{udandaro2024fomo_in_flux, magmax}. 
However, many existing routing and merging strategies rely on task-aware selection or assume static task boundaries, making them ill-suited for real-world retrieval;
this motivates the need for adaptive, task-agnostic routing and merging mechanisms for CL capable of navigating distribution shifts and cross-domain ambiguity.


\section{Background and Problem Formulation}
\label{sec:notation}

\vspace{-0.2cm}

\paragraph{Multimodal Dual-Encoder Models.}
We assess cross-modal similarity between image and text pairs using a multimodal embedding model. This model consists of an image encoder $f(\cdot; \theta)$ and a text encoder $g(\cdot; \phi)$. Given an input image $\mathbf{v}$ and its corresponding text $\mathbf{c}$, the model maps them to a shared latent space to produce embeddings $\mathbf{z} = f(\mathbf{v}; \theta)$ and $\mathbf{h} = g(\mathbf{c}; \phi)$. The alignment between the image and text is then evaluated using cosine similarity: $s(\mathbf{v}, \mathbf{c}) = \text{sim}(\mathbf{z}, \mathbf{h})$.

\vspace{-0.2cm}

\paragraph{Cross-Modal Retrieval.}
A goal of the retrieval system is to find the most relevant reference items from a gallery $\mathcal{R} = \{r_j\}_{j=1}^N$ given a query $q_i$ from a query set $\mathcal{Q} = \{q_i\}_{i=1}^N$. We assume an evaluation dataset with paired annotations, such that the unique ground-truth target for query $q_i$ is $r_i$.

The retrieval system computes the similarity score $s(q_i, r_j)$ between the query and all reference items in $\mathcal{R}$, ranking the references in descending order based on this score. Let $\text{rank}(q_i, r_i)$ denote the rank of the ground-truth reference $r_i$ among the sorted candidates. For a single query sample $q_i$, the Recall@K (R@k) metric acts as an indicator function, evaluating to $1$ if the correct reference is present within the top $K$ highest-ranked results, and $0$ otherwise. 
The overall retrieval performance on the dataset is then computed by averaging this single-sample indicator across all $N$ queries:
\vspace{-0.3cm}

\begin{equation}
    \text{Recall@K} = \frac{1}{N} \sum_{i=1}^N \mathbbm{1}[\text{rank}(q_i, r_i) \leq K].
\end{equation}

\vspace{-0.2cm}

Given the multimodal model described above, and an evaluation dataset $\mathcal{D} = \{(\mathbf{v}_i, \mathbf{c}_i)\}_{i=1}^N$, we can apply this general formulation to assess cross-modal alignment between images and texts in the dataset in both directions. For Image-to-Text~(I2T) retrieval, we treat images as the queries and texts as references~($\mathcal{Q} = \{\mathbf{v}_i\}_{i=1}^N$, $\mathcal{R} = \{\mathbf{c}_j\}_{j=1}^N$), and rank candidates according to $s(\mathbf{v}_i, \mathbf{c}_j)$. Conversely, for Text-to-Image~(T2I) retrieval, the text serves as the query against an image gallery~($\mathcal{Q} = \{\mathbf{c}_i\}_{i=1}^N$ and $\mathcal{R} = \{\mathbf{v}_j\}_{j=1}^N$), and ranking is performed based on $s(\mathbf{c}_i, \mathbf{v}_j)$.
\vspace{-0.2cm}

\paragraph{Continual Learning Setup.}
We further consider a multimodal dual-encoder model under a continual learning scenario, where it is trained on a sequence of $T$ tasks. We use the index $t \in \{1, \dots, T\}$ to denote a specific task, where $t=0$ refers to the pretrained state that serves as the initialization for our continual learning setup. 
For a given task $t$, the dataset consists of $N_t$ image-text pairs. We denote the $j$-th sample within this task as $(\mathbf{v}^{t}_{j}, \mathbf{c}^{t}_{j})$, where $j \in \{1, \dots, N_t\}$. Consequently, the parameters of the encoders after training on task $t$ are updated to $\theta^t$ and $\phi^t$, and we denote the models obtained at the end of each task as 
$    
f^t(\mathbf{v}) = f(\mathbf{v}; \theta^t),\ g^t(\mathbf{c}) = g(\mathbf{c}; \phi^t).
$

After learning task $t$, the model is evaluated on cross-modal retrieval across all previously seen tasks. 
To differentiate between the retrieval directions, let $m \in \{\text{I2T}, \text{T2I}\}$ denote the cross-modality direction and define $R^{m}_{t, k}$ as the R@K performance in direction $m$ on the evaluation dataset of a prior task $k$ (where $k \leq t$), computed using the updated encoders $f^t$ and $g^t$. To evaluate the overall performance at the end of the training sequence, we compute the Average Final R@K for each direction, which measures the model's retention across all $T$ tasks:
\vspace{-0.1cm}
\begin{equation}
    A^m_T = \frac{1}{T} \sum_{k=1}^T R^m_{T, k}.
    \label{eq:avg_final_recall}
\end{equation}
Similarly, we track the Average Incremental R@K by averaging the mean recall at every incremental step $t$. We provide the formal definition of this metric in \Cref{app:metric_definitions} alongside other supplementary metrics for evaluating continual learning systems.

\begin{table}[!t]
\centering
\caption{
Comparison of existing retrieval benchmarks for CL (see \Cref{app:summary_of_related_protocols}~for~more~discussion).
}
\label{tab:benchmark_comparison}
\small
\setlength{\tabcolsep}{6pt}
\renewcommand{\arraystretch}{1}

\begin{tabular}{lccccc}
\toprule
\textbf{Benchmark}
& \textbf{Diversity}
& \textbf{Specificity}
& \textbf{Curriculum}
& \textbf{OOD Robustness}
& \textbf{Accessibility} \\
\midrule

\citet{kai2021cross_modal_retrieval}~(2021)
& \xmarkr & \cmarkg & \xmarkr & \xmarkr & \cmarkg \\

\citet{ni2023off_diagonal}~(2023)
& \xmarkr & \xmarkr & \xmarkr & \xmarkr & \cmarkg \\

\citet{cui2024knowledge_rectification}~(2024)
& \cmarkg & \cmarkg & \xmarkr & \xmarkr & \cmarkg \\

\citet{garg2024tic_clip}~(2024)
& \cmarkg & \xmarkr & \cmarkg & \cmarkg & \xmarkr \\

\citet{c_clip2025}~(2025)
& \cmarkg & \xmarkr & \xmarkr & \cmarkg & \xmarkr \\

\midrule

\textbf{Our Framework}
& \cmarkg & \cmarkg & \cmarkg & \cmarkg & \cmarkg \\

\bottomrule

\vspace{-0.8cm}

\end{tabular}
\end{table}

\vspace{-0.2cm}

\section{A New Principled Framework for Continual Retrieval Evaluation}
\label{sec:framework}
\vspace{-0.2cm}

Standard benchmarks for multimodal CL often emphasize class-incremental proxies~(e.g., ImageNet-100) and neglect retrieval-specific requirements like global embedding consistency and fine-grained cross-modal alignment. To address this, we propose a new evaluation framework based on five core design choices to rigorously assess continual multimodal retrieval.

\vspace{-0.2cm}

\paragraph{Domain Diversity.}

To minimize overlap with pre-training data, we propose a sequence of heterogeneous domains: natural images~(\textbf{Flickr30K}~\citep{young-etal-2014-image}), AI-generated content~(\textbf{Lexica-SD}~\citep{yuwan0_lexica_stable_diffusion_v15}, \textbf{KreaM}~\citep{hahminlew_kream_product_blip_captions}), artwork~(\textbf{WikiArt}~\citep{atermors_wikiart_recaption}), specialized distributions like cartoons~(\textbf{Flintstones}~\citep{Kapuriya2025FlintstonesSVI,janak12_flintstonessvplusplus}), sketches~(\textbf{Sketch}~\citep{chowdhury2022fs,zoheb_sketch_scene}), and medical imaging~(\textbf{ROCOv2}~\citep{R_ckert_2024}). This diversity ensures realistic distribution shifts and prevents overly optimistic performance assessments.

\vspace{-0.2cm}

\paragraph{Semantic Granularity and Specificity.}
Many common datasets (e.g., Oxford Pets~\citep{parkhi2012cats} used by \citet{c_clip2025}) suffer from pair redundancy, where multiple images might match a generic caption or vice versa. Therefore, we select datasets with high semantic precision and distinct targets to ensure cross-modal retrieval metrics ($\text{I2T}$ and $\text{T2I}$) provide reliable performance signals.

\vspace{-0.2cm}

\paragraph{Difficulty-Calibrated Curriculum.} 
Since task ordering heavily impacts CL performance~\citep{Masana2020Class_orderings}, we use the frozen backbone's zero-shot performance as a proxy for task complexity, inspired by class-incremental works such as \citet{menabue2024semantic}. This creates a structured progression from in-distribution to challenging OOD domains, allowing us to evaluate plasticity degradation while providing a modular way to integrate future tasks.

\vspace{-0.2cm}

\paragraph{OOD Retrieval Generalization.} 
Robust methods should leverage sequentially acquired knowledge to enhance its generalization to entirely unseen domains. Therefore, we evaluate the model's capacity for OOD retrieval by assessing I2T and T2I performance on held-out, broad-domain benchmarks~(\textbf{COCO}~\citep{lin2014microsoft}, \textbf{NoCaps}~\citep{agrawal2019nocaps}) to quantify how effectively the model transfers learned cross-modal alignments to novel, open-vocabulary concepts beyond its training trajectory.

\vspace{-0.2cm}

\paragraph{Practicality, Accessibility, and Reproducibility.} 
Our framework simulates common practical fine-tuning scenarios and is designed for computational tractability for the broader community, and we fully open source our code to ensure standardized, reproducible comparisons.

We provide a detailed comparison of our framework against existing benchmark suites in \Cref{tab:benchmark_comparison}, highlighting how we address critical gaps in prior work. 
As demonstrated in \Cref{sec:experiments}, current CL strategies fail to provide consistent gains on this more challenging framework. 

\section{Dynamic Adapter Routing (DAR)}
\label{sec:method}

We propose Dynamic Adapter Routing (DAR), a novel continual multimodal learning method that operates on top of the pretrained dual-encoder architecture introduced in \Cref{sec:notation}. 
DAR learns a set of task-specific adapter modules and maintains a set of prototypes that anchor each task $t \in \{1, \dots, T\}$ in the shared backbone feature space. 
During inference, it routes each query to the most relevant adapter in a task-agnostic manner, using cosine similarity to the stored prototypes. 
To facilitate knowledge transfer for ambiguous samples, DAR employs dynamic adapter merging.

\vspace{-0.1cm}
\paragraph{Model and Adapter Design.}
We adopt the frozen pre-trained multimodal model with image encoder $f(\cdot; \theta^0)$ and text encoder $g(\cdot; \phi^0)$, where $\theta^0$ and $\phi^0$ denote the initial pretrained parameters (i.e., at state $t=0$). For each incoming task $t$, we introduce task-specific LoRA adapters, denoted as $\Delta\theta^t$ and $\Delta\phi^t$. Thus, the encoders adapted for task $t$ are formulated as:
\begin{equation}
f^t(\mathbf{v}) = f(\mathbf{v}; \theta^0 + \Delta\theta^t), \quad g^t(\mathbf{c}) = g(\mathbf{c}; \phi^0 + \Delta\phi^t)
\end{equation}

We apply LoRA to the attention output projection and the feed-forward networks (\texttt{attn.out\_proj}, \texttt{mlp.c\_fc}, and \texttt{mlp.c\_proj}), obtaining a set of lightweight, task-specific experts that share a common backbone representation. 
After training on task $t$, we freeze $\Delta\theta^t$ and $\Delta\phi^t$, and initialize the new adapters for the next task from the pretrained parameters $(\theta^0, \phi^0)$. 

To promote parameter reuse and avoid introducing unnecessary adapters, we reuse the adapters from previously seen similar tasks.
Specifically, given a new dataset, we compute its prototype using the frozen backbone features and compare it against the stored prototypes of previously learned adapters; if the similarity to an existing prototype exceeds a predefined threshold, we reuse the corresponding adapter and continue training its LoRA parameters, rather than initializing a new adapter from scratch.

\vspace{-0.1cm}
\paragraph{Prototype Memory and Margin of Similarity Score.}
To enable task-agnostic routing, we summarize each task $t$ using prototype vectors in the shared latent space. After training on dataset $\mathcal{D}_t = \{(\mathbf{v}^{t}_{j}, \mathbf{c}^{t}_{j})\}_{j=1}^{N_t}$, we compute an image prototype $\mathbf{p}_{\mathbf{v}}^t$ and a text prototype $\mathbf{p}_{\mathbf{c}}^t$ by averaging the normalized backbone embeddings over the task dataset.
At inference time, given an image-text pair $(\mathbf{v}, \mathbf{c})$, we obtain normalized backbone embeddings $\mathbf{z}$ and $\mathbf{h}$. 
We then compare the image embedding against the image prototypes and the text embedding against the text prototypes to obtain the routing score $S^t$ corresponding to the prototype from the task $t$:
\vspace{-0.2cm}
\begin{equation}
S^t_{\mathbf{v}} = \text{sim}(\mathbf{z}, \mathbf{p}_{\mathbf{v}}^t), \quad
S^t_{\mathbf{c}} = \text{sim}(\mathbf{h}, \mathbf{p}_{\mathbf{c}}^t), \quad
S^t = \frac{1}{2}\left(S^t_{\mathbf{v}} + S^t_{\mathbf{c}}\right).
\end{equation}
We then take these scores to identify the top-$1$ and top-$2$ similarities, denoted as $S^{(1)}$ and $S^{(2)}$, and compute their similarity margin $M$ as 
$
M = S^{(1)} - S^{(2)}.
$
If $M$ falls below a set threshold $\gamma$, we interpret this as ambiguity between tasks, and perform the adapter merging.

\vspace{-0.1cm}
\paragraph{Continual Adapter Merging.}
When ambiguity is detected ($M < \gamma$), we apply adaptive merging. For a given input query $q$, we compute merging weights $w^t$ for the top-$2$ adapters using a temperature-scaled softmax over their similarity scores $S^t$, and merge the adapter parameters as follows:
\begin{equation}
w^t = \frac{\exp(S^t / \tau)}{\sum_{k \in \text{top-}2} \exp(S^k / \tau)} , \quad
\Delta\theta^* = \sum_{t \in \text{top-}2} w^t \Delta\theta^t, \quad
\Delta\phi^* = \sum_{t \in \text{top-}2} w^t \Delta\phi^t.
\end{equation}
By smoothly interpolating between task-specific adapters in such adaptive way, our method improves robustness to distribution shifts and effectively navigates cross-domain ambiguity without increasing the inference-time parameter count, yielding both in-domain stability and generalization.

\section{Experiments}
\label{sec:experiments}

We use framework described in \Cref{sec:framework} 
and evaluate
\textbf{DAR}~(\Cref{sec:method}) method with a diverse set of CL strategies for VLMs, which includes regularization-based and parameter-efficient approaches. We employ simple baselines, such as \textbf{Zero-shot~(ZS)} performance of the backbone and \textbf{Fine-tuning~(FT)}, where the model is sequentially updated without explicit mechanisms to prevent forgetting. Moreover, we include standard CL approaches applicable to retrieval models, such as \textbf{EWC}~\citep{kirkpatrick2017ewc} and \textbf{L2P}~\citep{Wang2021LearningTP} and \textbf{Task Arithmetic~(TA)}~\citep{ilharco2023editing}. Finally we also evaluate methods that were created directly for retrieval task such as \textbf{Mod-X}~\citep{ni2023off_diagonal}, \textbf{DKR}~\citep{cui2024knowledge_rectification} and \textbf{C-CLIP}~\citep{c_clip2025}.

Unless explicitly stated otherwise, we build our CL setup on top of a frozen CLIP ViT-B/16 backbone, and report Average Final Recall@1 as described in \Cref{eq:avg_final_recall} to measure cross-modal retrieval performance. 
All methods are trained using AdamW for 20 epochs with a batch size of $256$ and learning rate $10^{-4}$ .
We implement DAR using task-specific LoRA adapters inserted into both the visual and text encoders. 
Unless otherwise specified, we use rank r=$8$, top-k=$2$ routing, and trigger adaptive merging when the similarity margin falls below $\gamma$=$0.05$.
We also use CoreSpace~\citep{panariello2026merging_corespace} merging. 
We provide the full implementation details in \Cref{app:implementation_details} and share the exact hyperparameters for our runs in the open-sourced code.

\begin{table*}[!t]
  \centering
  \caption{
  Cross-modal \textbf{Recall@1} at the end of continual training on our proposed evaluation framework for CLIP ViT-B/16. Surprisingly, commonly used CL approaches often fail to improve upon fine-tuning baseline, which highlights how continual retrieval presents its own set of challenges that cannot be addressed simply by adapting CIL methods. Our novel method, \textbf{DAR}, shows robust performance and outperforms the previous approaches by a sizable margin.
  }
  \label{tab:main_table}
  \resizebox{\linewidth}{!}{
  \setlength{\tabcolsep}{2pt}
  \begin{tabular}{l ccccccc >{\columncolor{gray!15}}c ccccccc >{\columncolor{gray!15}}c}
  \toprule
  & \multicolumn{8}{c}{\textbf{Text $\rightarrow$ Image}} & \multicolumn{8}{c}{\textbf{Image $\rightarrow$ Text}} \\
  \cmidrule(lr){2-9} \cmidrule(lr){10-17}
  & \textbf{\footnotesize Flickr}  
  & \textbf{\footnotesize Lexica}  
  & \textbf{\footnotesize WikiArt}  
  & \textbf{\footnotesize KreaM}  
  & \textbf{\footnotesize Flints}  
  & \textbf{\footnotesize Sketch}  
  & \textbf{\footnotesize ROCO}  
  & \textbf{\footnotesize Avg.}  
  & \textbf{\footnotesize Flickr}  
  & \textbf{\footnotesize Lexica}  
  & \textbf{\footnotesize WikiArt}  
  & \textbf{\footnotesize KreaM}  
  & \textbf{\footnotesize Flints}  
  & \textbf{\footnotesize Sketch}  
  & \textbf{\footnotesize ROCO}  
  & \textbf{\footnotesize Avg.} \\
  \midrule
  
  \textbf{ZS}  
  & 62.3 & 52.3 & 22.6 & 20.0 & 16.6 & 5.2 & 1.8 & 25.8  
  & 82.0 & 52.0 & 20.8 & 20.2 & 11.1 & 4.2 & 1.5 & 27.4 \\
  
  \textbf{FT}  
  & 73.5 & 63.0 & 37.1 & 26.7 & 38.3 & 8.3 & 6.5 & 36.2  
  & 88.5 & 64.8 & 38.5 & 28.4 & 35.4 & 8.5 & 6.9 & 38.7 \\
    
  \textbf{TA}  
  & 73.2 & 62.5 & 35.0 & 24.1 & 32.8 & 8.2 & 3.2 & 34.1  
  & 88.7 & 64.9 & 35.2 & 24.4 & 27.5 & 7.8 & 3.6 & 36.0 \\
  
  \textbf{EWC}  
  & 75.4 & 62.0 & 36.3 & 31.6 & 42.3 & 12.3 & 8.6 & 38.4  
  & 89.5 & 62.5 & 36.5 & 33.3 & 38.8 & 12.4 & 8.6 & 40.2 \\
  
  \textbf{Mod-X}  
  & 73.5 & 61.0 & 36.4 & 27.6 & 40.1 & 9.4 & 9.4 & 36.8  
  & 88.5 & 60.9 & 36.8 & 28.9 & 36.9 & 8.6 & 9.3 & 38.5 \\
  
  \textbf{C-CLIP}  
  & 73.3 & 61.9 & 34.3 & 25.1 & 32.6 & 8.9 & 3.7 & 34.3  
  & 88.3 & 65.7 & 34.1 & 26.1 & 26.7 & 7.2 & 3.3 & 35.9 \\
  
  \textbf{L2P}  
  & 66.3 & 51.5 & 24.3 & 18.4 & 20.7 & 6.0 & 2.9 & 27.2
  & 83.2 & 51.1 & 20.6 & 14.9 & 15.5 & 4.2 & 2.5 & 27.4 \\

  \textbf{DKR}  
  & 62.9 & 45.0 & 24.7 & 24.5 & 36.6 & 9.9 & 12.5 & 30.9
  & 78.0 & 38.1 & 20.4 & 22.1 & 33.1 & 7.1 & 12.2 & 30.1 \\

  \textbf{DAR} 
  & \textbf{81.0} & \textbf{77.9} & \textbf{50.2} & \textbf{45.1} & \textbf{49.6} & \textbf{15.8} & \textbf{13.2} & \textbf{47.5} 
  & \textbf{93.4} & \textbf{77.4} & \textbf{51.3} & \textbf{45.3} & \textbf{48.2} & \textbf{16.2} & \textbf{13.3} & \textbf{49.3} \\
  
  \bottomrule
  \end{tabular}
  }
\end{table*}

\begin{figure}[t]
    \centering
    \includegraphics[width=0.49\linewidth]{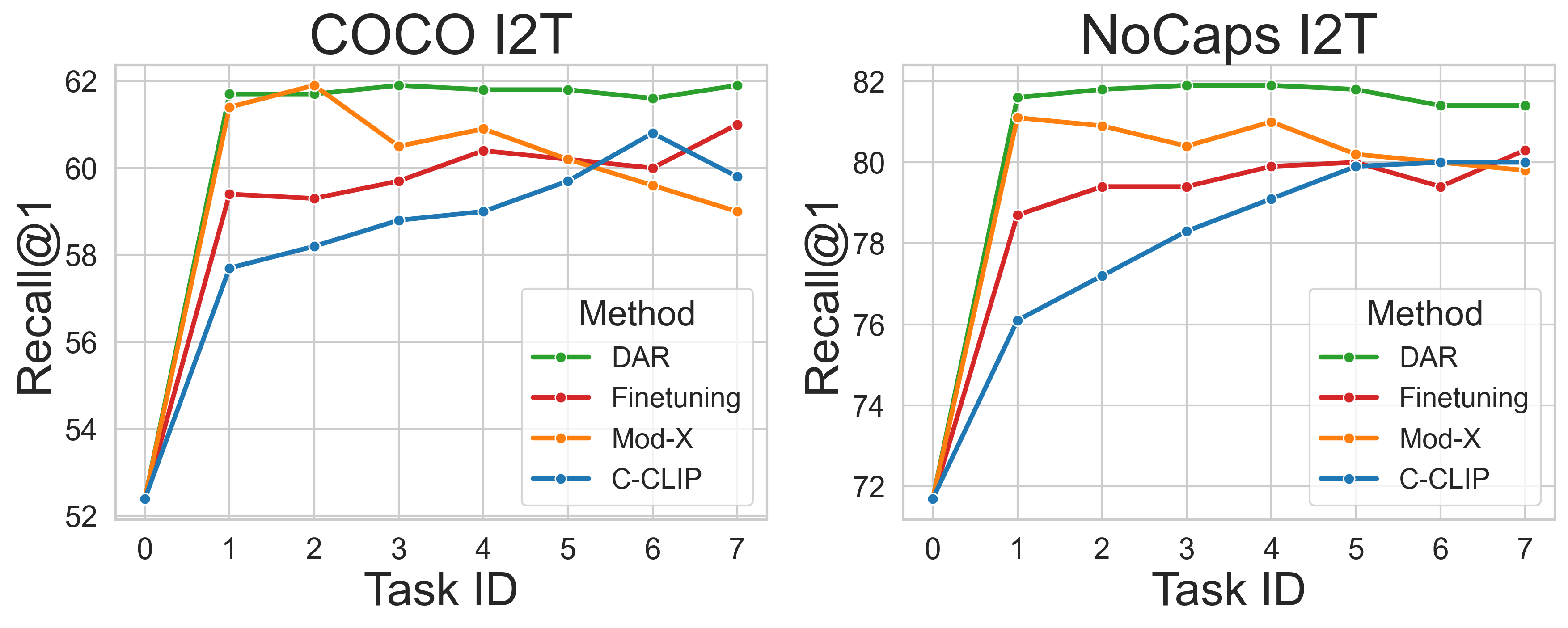}
    \includegraphics[width=0.49\linewidth]{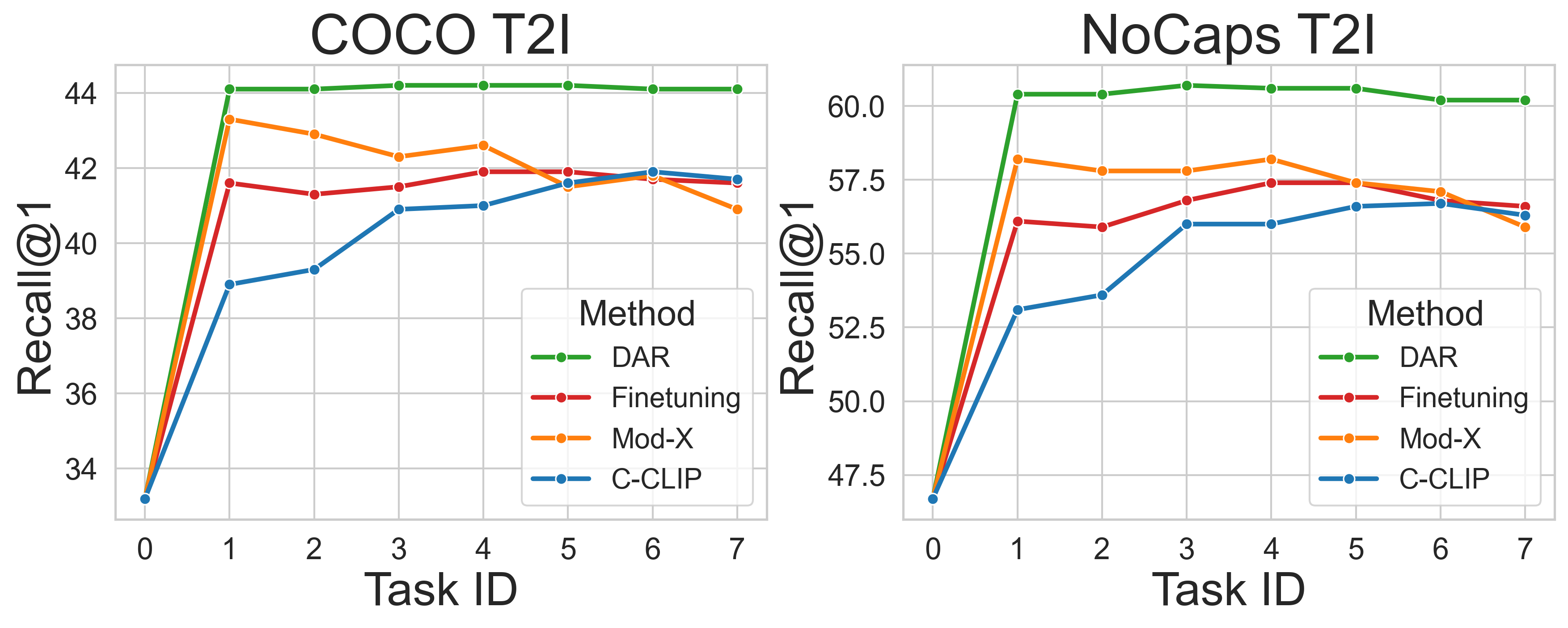}

    \caption{
    (left) Image-to-Text and (right) Text-to-Image Recall@1 performance of various CL methods evaluated on the COCO and NoCaps datasets. \textbf{DAR} consistently outperforms alternative approaches on the held-out datasets throughout continual training on the task suite from \Cref{tab:main_table}, which highlights the efficacy of knowledge-sharing mechanisms in our method.
    }
    \label{fig:ood_r_main}
\end{figure}

\subsection{Main Results}
\label{sec:main_results}

In \Cref{tab:main_table}, we report the Recall@1 for both I2T and T2I retrieval across several CL methods evaluated within our framework. 
Interestingly, naive fine-tuning emerges as a strong baseline under our evaluation settings, outperforming or closely matching dedicated CL methods such as C-CLIP and Mod-X. 
Overall, our results demonstrate that existing methods fail to consistently address the challenges of continual retrieval. 
In contrast, DAR provides a substantial improvement over the baseline, outperforming fine-tuning by approximately 11\% for both I2T and T2I. 
This performance gap widens further when we measure Recall@5 in \Cref{app:additional_experiments}, underscoring the efficacy of an approach designed specifically for retrieval. Additionally, we evaluate zero-shot retrieval performance on the held-out COCO and NoCaps datasets in \Cref{fig:ood_r_main}. 
DAR consistently achieves the strongest results, demonstrating that, despite continual updates, our method preserves and even improves the global alignment structure required for generalization in retrieval tasks. 
We provide numerical results for these experiments in \Cref{app:full_zero_shot}.

\clearpage

\begin{wraptable}[7]{r}{0.55\textwidth}
\centering
\vspace{-1.4cm}
\caption{DAR and VLM backbones.}
\label{tab:backbones}
\small
\begin{tabular}{lcccccc}
\toprule
\multirow{2}{*}{Method} & \multicolumn{2}{c}{ViT-B/32} & \multicolumn{2}{c}{ViT-B/16} & \multicolumn{2}{c}{ViT-L/14} \\
\cmidrule(lr){2-3} \cmidrule(lr){4-5} \cmidrule(lr){6-7}
 & T$\rightarrow$I & I$\rightarrow$T  & T$\rightarrow$I & I$\rightarrow$T  & T$\rightarrow$I & I$\rightarrow$T \\ \midrule
FT & 33.4 & 31.1 & 36.2 & 38.7 & 44.0 & 45.5 \\
Mod-X & 32.9 & 30.6 & 36.8 & 38.5 & 44.0 & 45.8 \\
C-CLIP &  32.1 & 29.2 & 34.3 & 35.9 & 41.0 & 42.3 \\
\textbf{DAR} & \textbf{42.3} & \textbf{43.7} & \textbf{47.5} & \textbf{49.3} & \textbf{52.8} & \textbf{53.7} \\ \bottomrule
\end{tabular}
\end{wraptable}

\subsection{VLM Backbone Generalization}
\label{sec:backbones}

To assess whether our proposed method generalizes across model scales and representation capacities, we evaluate CL methods using different CLIP backbones under the same protocol as in \Cref{sec:main_results}. 
Specifically, we compare ViT-B/32, ViT-B/16, and ViT-L/14, providing additional results with our method applied to larger and smaller models. 
We present the results for DAR and the best-performing CL baselines in \Cref{tab:backbones}. 
As expected, stronger backbones improve retrieval performance for all methods. 
However, DAR consistently outperforms the competing approaches across all scales, demonstrating the general applicability and robustness of our method.

\subsection{DAR Robustness on Alternative Task Sequences}

\begin{wrapfigure}[17]{r}{0.4\textwidth}
    \centering
    \vspace{-0.4cm}
    \includegraphics[width=\linewidth]{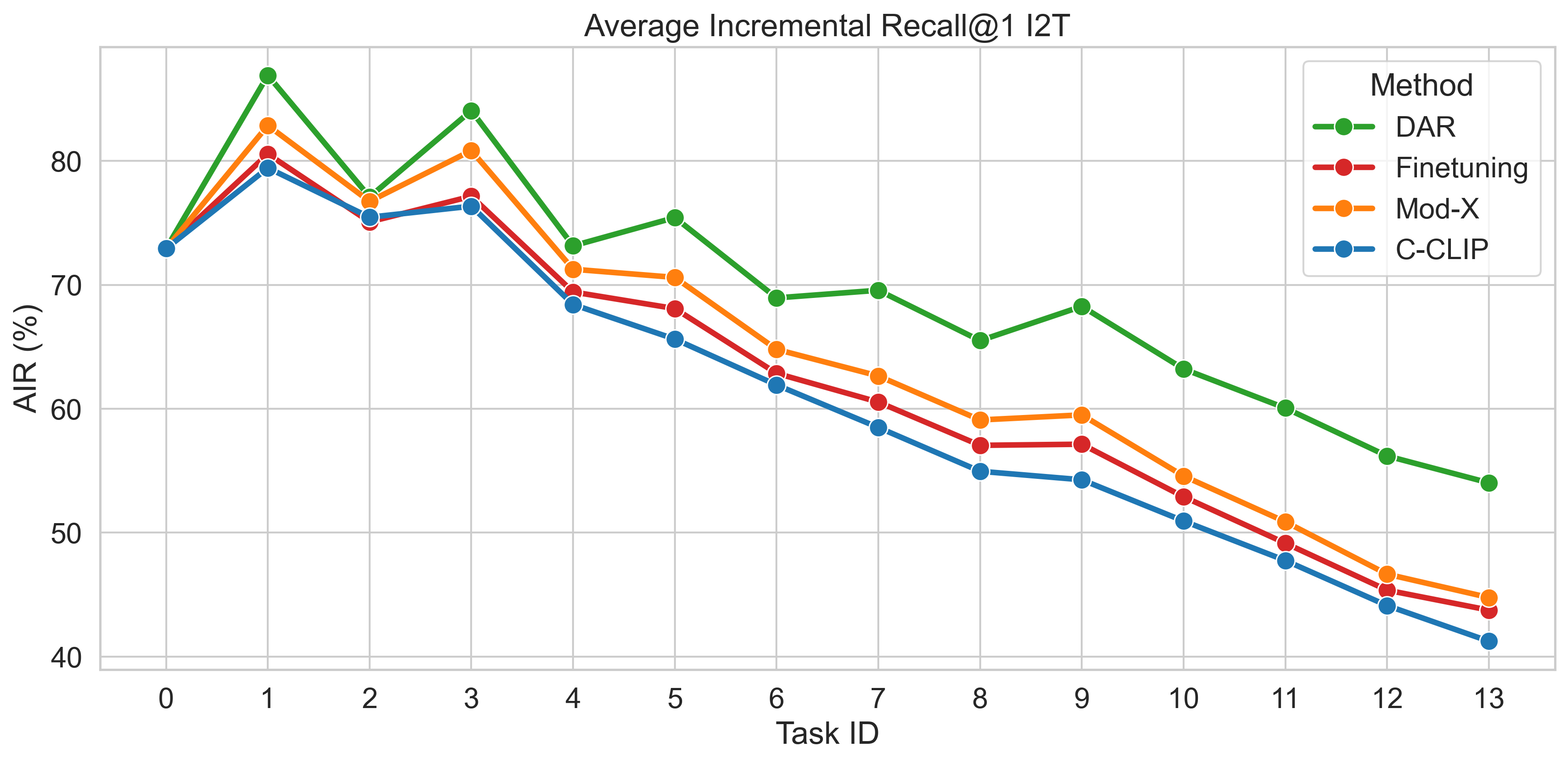}
    
    
    \includegraphics[width=\linewidth]{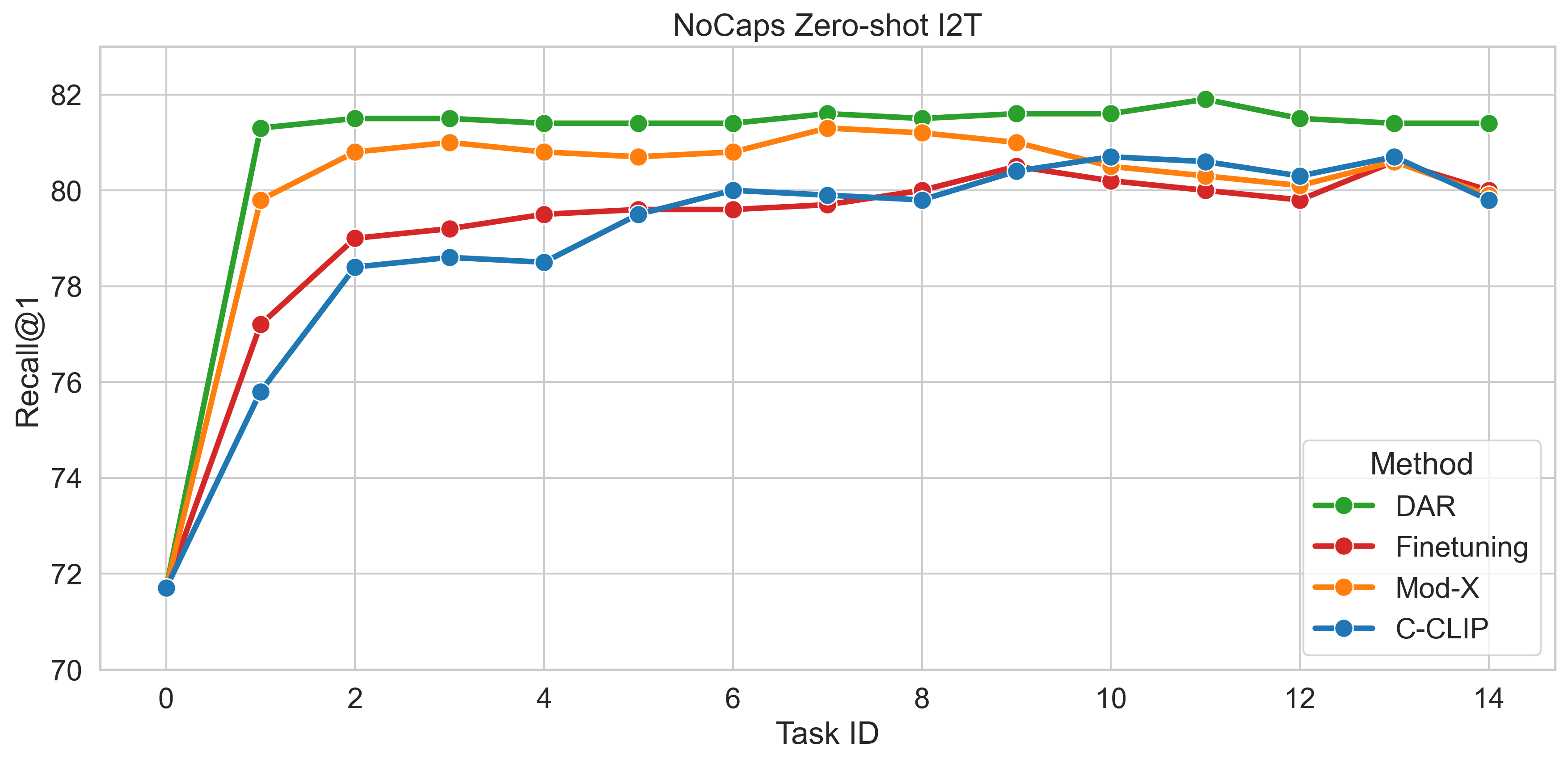}

    \caption{(top) In-distribution AIR and (bottom) I2T R@1 on NoCaps during prolonged continual training.}
    \label{fig:ood_seq_14}
\end{wrapfigure}

To evaluate DAR and its adapter sharing mechanism, we extend the training sequences by splitting each task from \Cref{tab:main_table} into two consecutive halves that are fed sequentially (e.g., Flickr[:50\%], Flickr[50\%:], etc.). 
This modified framework tests the modularity of our method and its stability over longer horizons, where representation drift and routing ambiguity accumulate. 
In \Cref{fig:ood_seq_14}, we show the Average Incremental Recall and Image-to-Text Recall@1 on held-out NoCaps dataset throughout this extended sequence.

Even in this more challenging setting, DAR remains the top-performing CL method. 
Crucially, the adapter sharing mechanism functions as intended, efficiently allocating a single adapter for both corresponding halves of a dataset. 
This highlights our method's scalability, robustness, and broad applicability.
In \Cref{app:sec_full_results_long_sequences}, we examine more task sequences, and show that our method consistently proves to be the most effective under diverse CL protocols.

\subsection{Retrieval Performance Analysis and Correlation with Cross-modal Alignment}

\begin{wrapfigure}[8]{r}{0.5\textwidth}
    \centering
    \vspace{-0.6cm}
    \includegraphics[width=\linewidth]{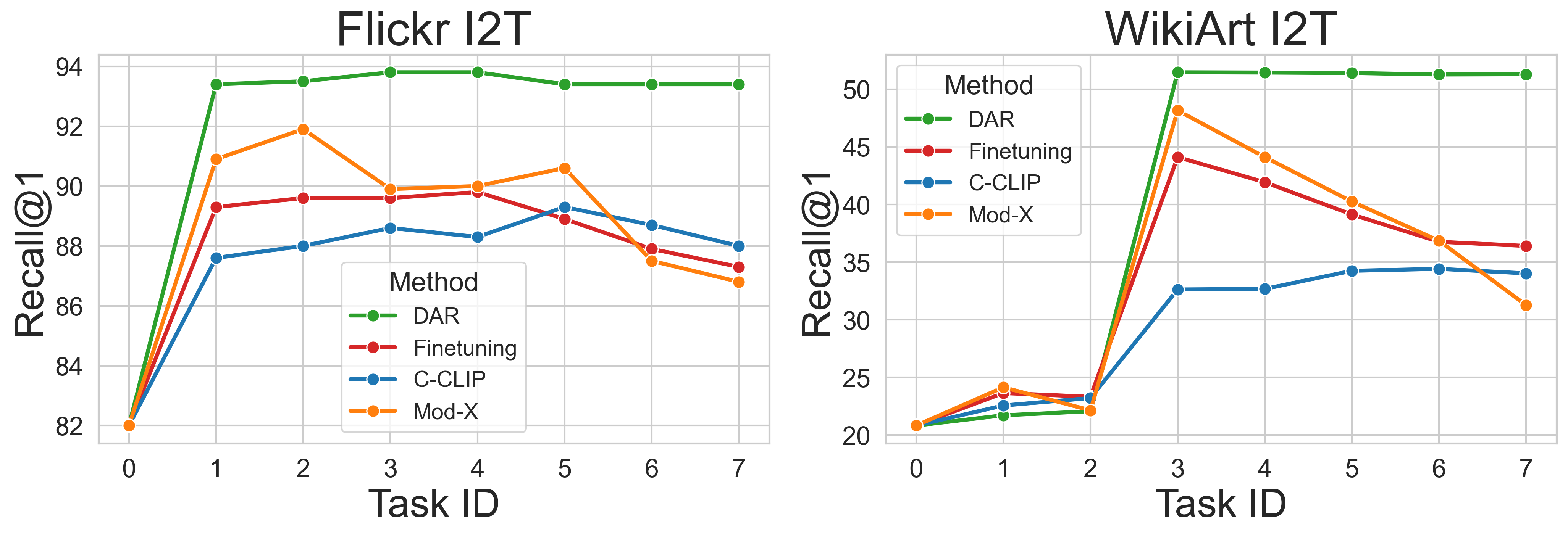}
    \caption{Continual performance on selected tasks.}
    \label{fig:i2t_flickr_wikiart}
\end{wrapfigure}

To understand what drives the superior performance of DAR, we take a closer look at the stability-plasticity trade-off inherent in our method. 
In \Cref{fig:i2t_flickr_wikiart}, we compare its performance on the first and third training tasks throughout the continual learning process against selected  baselines.
Interestingly, we find that the superior performance of our method is not solely driven by its ability to separate tasks, which provides unmatched stability.
Our LoRA adapters also exhibit excellent plasticity, enabling the model to effectively learn new incoming tasks and achieve the highest performance on a given task immediately after learning it.
The enhanced plasticity is likely a result of absence of regularization employed by other methods, and because we initialize the adapter training for each task from a more general, plastic backbone.

\begin{table}[t]
\centering
\setlength{\tabcolsep}{2pt}
\caption{
Final retrieval and cross-modal alignment metrics for selected CL methods evaluated on our benchmark. Arrows indicate whether a higher ($\uparrow$) or lower ($\downarrow$) metric is desirable. 
We observe a strong correlation between modality alignment and retrieval performance.
DAR achieves superior results by most effectively preserving this alignment throughout the training process.
}
\label{tab:additional_measures}
\small
\begin{tabular}{lccccccccc}
\toprule

& \multicolumn{3}{c}{Retrieval} 
& \multicolumn{1}{c}{Cls.}
& \multicolumn{3}{c}{Local Alignment} 
& \multicolumn{2}{c}{Structure} \\
\cmidrule(lr){2-4} \cmidrule(lr){5-5} \cmidrule(lr){6-8} \cmidrule(lr){9-10}
Method 
& R@1 $\uparrow$ & R@5 $\uparrow$ & MRR $\uparrow$
& Acc. $\uparrow$
& \begin{tabular}{@{}c@{}}Negative\\Margin $\uparrow$\end{tabular} 
& \begin{tabular}{@{}c@{}}Violation\\Rate $\downarrow$\end{tabular} 
& \begin{tabular}{@{}c@{}}Asymmetry\\Gap $\downarrow$\end{tabular}
& \begin{tabular}{@{}c@{}}Neighbour\\Overlap $\uparrow$\end{tabular} 
& CKA $\uparrow$ \\
\midrule
ZS     & 26.6 & 43.2 & 0.3 & 61.8 & 0.00 & 0.74 & 4.89 & -- & 0.42 \\
FT     & 37.5 & 57.8 & 0.4 & 62.3 & 0.01 & 0.64 & 1.58 & 0.60 & 0.56 \\
C-CLIP & 35.1 & 53.6 & 0.39 & \textbf{63.5} & 0.01 & 0.66 & 2.73 & 0.80 & 0.52 \\
Mod-X  & 37.7 & 59.0 & 0.41 & 59.1 & 0.01 & 0.65 & 2.19 & 0.52 & 0.55 \\
\textbf{DAR} & \textbf{48.4} & \textbf{71.3} & \textbf{0.50} & 58.5 & \textbf{0.03} & \textbf{0.52} & \textbf{1.17} & \textbf{0.86} & \textbf{0.6} \\
\bottomrule
\end{tabular}%
\vspace{-0.5cm}
\end{table}

To analyse the performance of continual retrieval more deeply, we evaluate selected CL methods after training on our benchmark in \Cref{tab:additional_measures}. We report retrieval with additional classification, and alignment metrics averaged across all training tasks.
Specifically, we measure symmetric R@1, R@5, and MRR (averaged over I2T and T2I), as well as mean zero-shot accuracy across ImageNet, CIFAR, EuroSAT, and DomainNet. 
Additionally, we quantify local cross-modal separation using the Top-10 negative margin, hard-negative violation rate, and cross-modal asymmetry gap. 
Finally, we assess the representation stability and structural integrity via top-10 neighbor overlap between the image and text embeddings and linear CKA. See \Cref{app:metric_definitions} for the full definitions of the metrics.

Our analysis reveals how retrieval performance is fundamentally tied to embedding geometry. DAR achieves state-of-the-art continual retrieval by simultaneously improving local alignment and preserving cross-modal structure. However, there is also a distinct classification-retrieval trade-off: DAR's specialization in fine-grained retrieval inadvertently compromises the broader generalization required for robust zero-shot classification, causing it to underperform compared to methods that were designed with CIL in mind. We present detailed classification results in Appendix~\ref{app:full_classification_results}.

\vspace{-0.3cm}
\subsection{Low-Rank Adaptation and Memory Cost}

\begin{wraptable}[8]{r}{0.35\textwidth}
\vspace{-1.3cm}
\centering
\small
\caption{LoRA rank ablation.}
\label{tab:effect_of_adapter_capacity}
\begin{tabular}{lcc}
\toprule
 & T$\rightarrow$I & I$\rightarrow$T \\
\midrule
Full-FT & 36.2 & 38.7 \\
LoRA-FT (r=16) &  36.0 &  34.2  \\
\midrule
DAR - LoRA (r=4) & 46.9 & 48.3   \\
\textbf{DAR - LoRA (r=8)} & 47.5  & 49.3\\
DAR - LoRA (r=16) & 47.0 & 48.5 \\
DAR - LoRA (r=32) &  47.1 & 48.5\\
\bottomrule
\end{tabular}
\end{wraptable}

Because DAR relies on low-rank adapters, it is crucial to isolate whether our performance gains stem from the low-rank adaptation itself or from our method design and to which degree the adapter hyperparameters influence our results. 
To this end, we evaluate standard FT against fine-tuning conducted with LoRA; in both cases, the models use a single parameter set through the learning process, differing by either full or low rank updates.
Furthermore, since CL might span multiple tasks, we investigate the LoRA's inherent expressivity-compression trade-off to determine how much additional memory we require to successfully perform in practice.
To address these points, in \Cref{tab:effect_of_adapter_capacity} we compare DAR across various LoRA ranks against both Full-FT and LoRA fine-tuning baselines.

Standard LoRA fine-tuning does not outperform full fine-tuning, which confirms that our observed performance improvements are not simply an artifact of the underlying parameter-efficient fine-tuning (PEFT) method, but rather a direct result of DAR's design.
Crucially, DAR also maintains robust performance across all evaluated LoRA ranks and performs well even under aggressive parameter compression (e.g., $r=4$).
This indicates that the proposed routing and merging mechanisms can function effectively even with small adapters, allowing our method to scale seamlessly as the number of prototypes increases.
In practice, assuming standard model dimensions of $d_m=768$ and $d_h=3072$ with a rank of $r=8$, DAR adapters require only $2r(d_m + d_h)$ parameters per MLP layer and $2rd_m$ parameters for the output projection in attention. 
This yields an adapter-to-full-weight parameter ratio of approximately $\frac{r}{d_m}$ per block, meaning we can store roughly 96 task-specific adapters using the exact same memory footprint as a single copy of the full model weights. 
DAR's actual memory footprint is also reduced even by adapter reuse via prototype routing.
As a result, the memory overhead of our method is substantially lower than that of methods like C-CLIP, Mod-X, or EWC, which store full model copies, unless we approach large numbers of adapters.

\subsection{Routing and Merging Strategies Ablation}

\begin{wraptable}[8]{r}{0.35\textwidth}
\vspace{-1.5cm}
\centering
\small
\caption{DAR routing strategies.}
\label{tab:routing_ablation}
\begin{tabular}{lcc}
\toprule
 & T$\rightarrow$I & I$\rightarrow$T  \\
\midrule
Random & 36.7 & 32.7 \\
Oracle & 47.1 & 48.6 \\
\midrule
Image-only & 46.9 & 48.4 \\
Text-only & 43.0 & 45.0 \\
Image+Text (max) & 46.6 & 48.2\\
Image+Text (avg) & 46.9 & 48.5 \\
\midrule
\textbf{DAR} & \textbf{47.5} & \textbf{49.3} \\
\bottomrule
\end{tabular}
\end{wraptable}

Next, we evaluate the impact of the hyperparameters and design of various DAR components. 
We first compare different prototype signals and fusion strategies that can be used to select task-specific adapters during inference.
We present the results for this ablation in \Cref{tab:routing_ablation}. 
We find that combining image and text similarities consistently yields better performance than unimodal routing. 
Moreover, averaging the similarities from both modalities even outperforms oracle routing based on ground-truth task labels,
which highlights how our routing mechanism enables cross-task knowledge transfer and validates our final design choice for DAR routing.

\begin{wraptable}[12]{r}{0.4\textwidth}
\centering
\vspace{-0.6cm}
\caption{Adaptive merging ablation.}
\label{tab:merging_strategy_combined}
\small
\begin{tabular}{cccccc}
\toprule
 & & \multicolumn{2}{c}{ID} & \multicolumn{2}{c}{OOD} \\
\cmidrule(lr){3-4} \cmidrule(lr){5-6}
$k$ & $\gamma$ & T$\rightarrow$I & I$\rightarrow$T & T$\rightarrow$I & I$\rightarrow$T \\
\midrule
1 & -- & 46.8 & 48.6 & 51.8 & 71.3 \\
2 & 0.00 & 46.9 & 48.5 & 52.0 & 71.2 \\
2 & 0.01 & 46.9 & 48.5 & 52.0 & 71.2 \\
2 & 0.05 & \textbf{47.5} & \textbf{49.3} & 52.2 & 71.4 \\
2 & 0.10 & 46.8 & 48.4 & \textbf{52.4} & 71.6 \\
3 & 0.05 & 46.9 & 48.3 & 52.3 & 71.5 \\
4 & 0.05 & 46.9 & 48.3 & 52.3 & \textbf{71.7} \\
\bottomrule
\end{tabular}
\end{wraptable}

Next, we ablate our adaptive merging mechanism in \Cref{tab:merging_strategy_combined} to assess how the number of merged adapters~($k$) and routing ambiguity threshold~($\gamma$) impact In-Distribution~(ID) and Out-of-Distribution (OOD)~retrieval reported as R@1 averaged across training and held-out tasks at the end of continual training. 
Overall, adaptive merging improves performance by combining information across tasks for ambiguous samples, and especially helps to improve OOD performance. 
However, effective merging requires sufficiently set routing uncertainty threshold: small values yield negligible gains over the no-merging baseline, and overly conservative merging with larger $\gamma$ degrades ID performance.
We highlight that, despite these trade-offs, DAR overall remains highly robust to hyperparameter variations.

\begin{wraptable}{r}{0.35\textwidth}
\centering
\vspace{-0.4cm}
\caption{Model merging ablation.}
\label{tab:merging_algorithms}
\small
\begin{tabular}{lcc}
\toprule
 &  T$\rightarrow$I & I$\rightarrow$T \\
\midrule
No merging & 46.8 & 48.6 \\
Avg (uniform) & 46.8 & 48.2 \\
TA~\citep{ilharco2023editing} & 46.9 & 48.4 \\
DARE-TIES~\citep{yadav2023ties} & 46.8 & 48.3 \\
ISO-C~\cite{marczak2025no} & 46.9 & 48.4 \\
CoreSpace~\cite{panariello2026merging_corespace} & \textbf{47.5} & \textbf{49.3} \\
\bottomrule
\end{tabular}
\end{wraptable}

While DAR employs CoreSpace~\cite{panariello2026merging_corespace} merging, its modular design easily supports alternative strategies to combine the routed adapters. 
In \Cref{tab:merging_algorithms}, we compare several representative merging methods using a fixed top-$2$ routing configuration and report the final R@1 averaged across training tasks. 

Methods designed for full-rank weight merging (Task Arithmetic, DARE-TIES, ISO-C) provide only marginal gains over naive averaging when applied to LoRA adapters. CoreSpace performs best and operates directly in the low-rank subspace of the adapters, aligning shared dominant directions while mitigating destructive interference between unrelated updates.

\section{Conclusions}
\label{sec:conclusions}

In this work, we introduce a principled benchmark for continual multimodal retrieval, establishing how continual retrieval poses fundamentally different challenges than continual classification. 
By evaluating existing CL methods on our benchmark, we reveal that their design is heavily skewed towards class-incremental settings, leading to suboptimal performance in this more demanding retrieval context. 
To bridge this gap, we propose Dynamic Adapter Routing (DAR), a novel method that employs lightweight adapters to significantly improve both retrieval performance and robustness across diverse domains. 
Ultimately, our findings underscore that continual retrieval is a distinct problem, necessitating tailored evaluation protocols and adaptation strategies.

\paragraph{Limitations and future work.}

Our routing mechanism relies on prototype similarity and the frozen backbone, which may become unreliable when domains strongly overlap or change drastically over time, potentially leading to degraded decisions. 
The current adaptive merging strategy also depends on manually selected hyperparameters, and more sophisticated strategies could further improve its robustness.
Nonetheless, our ablations suggest that DAR is quite robust to the hyperparameter choice.
While our evaluation focuses on staged CL with CLIP-style models, extending it to fully online settings and other multimodal architectures could be an important area for future research.

\paragraph{Broader impact.} Our work aims to advance the field of continual multimodal retrieval through robust evaluation protocols and novel techniques. 
Although we acknowledge the broader risks of machine learning misuse and advocate for the responsible development of this technology, we do not see any negative societal impacts specific to our research that would require explicit mention here.

\clearpage

\bibliographystyle{plainnat}
\bibliography{bibliography}

\clearpage
\appendix
\onecolumn

\Huge
\textbf{Appendix}
\normalsize

\section{Implementation details}
\label{app:implementation_details}

\paragraph{Global training protocol.}
Unless otherwise stated, all methods are trained with the same optimization,
data, and evaluation protocol. We use CLIP ViT-B/16 initialized from the
pretrained checkpoint, train for 20 epochs per task, and use a training batch
size of 256. Validation and zero-shot evaluation use batch size 512.  We use 32 data-loader workers and a maximum text length of 77 tokens. Retrieval evaluation is performed on the same task sequence and validation datasets for all methods.

\begin{table}[H]
\centering
\caption{Global hyperparameters shared by all methods.}
\begin{tabular}{ll}
\toprule
Hyperparameter & Value \\
\midrule
Backbone & CLIP ViT-B/16 \\
Epochs per task & 20 \\
Training batch size & 256 \\
Validation batch size & 512 \\
Zero-shot batch size & 512 \\
Max text length & 77 \\
Number of workers & 32 \\
Seed & 42 \\
\bottomrule
\end{tabular}
\end{table}

\paragraph{DAR-specific hyperparameters.}
DAR uses task-specific LoRA adapters inserted into both image and text encoders.
Unless otherwise stated, we use LoRA rank $r=8$, LoRA scaling $\alpha=16$, and
dropout 0.0. At inference time, we route using the average image-text prototype similarity,
select the top-2 adapters, and perform adaptive merging when the similarity
margin is below $\gamma=0.05$.

\begin{table}[H]
\centering
\caption{DAR-specific hyperparameters.}
\begin{tabular}{ll}
\toprule
Hyperparameter & Value \\
\midrule
Optimizer & AdamW \\
Learning rate & $1\times10^{-4}$ \\
Weight decay & 0.2 \\
Learning-rate scheduler & Cosine decay \\
Warmup epochs & 5 \\
LoRA rank & 8 \\
LoRA alpha & 16 \\
LoRA dropout & 0.0 \\
Train logit scale & False \\
Prototype momentum & 0.0 \\
Normalize prototypes & True \\
Routing top-$k$ & 2 \\
Margin threshold $\gamma$ & 0.05 \\
Evaluation adapter policy & Margin-based \\
\bottomrule
\end{tabular}
\end{table}

\section{Compute resources}
\label{app:compute_resources}

All experiments were conducted on a single NVIDIA A100 GPU with 40GB memory. Training each method on the main continual retrieval benchmark with 7 tasks required approximately 5-6 GPU hours per run, while the extended CL sequences consisting of 14 tasks required between 7-11 GPU hours depending on the method. Unless otherwise stated, all methods used the same training protocol described in \Cref{app:implementation_details}, including 20 epochs per task, training batch size 256, and the CLIP ViT-B/16 backbone.

\clearpage
\section{Additional metric definitions}
\label{app:metric_definitions}

Following the notation established in \Cref{sec:notation}, we define additional metrics helpful for assessing the performance of continual retrieval models.

To capture the model's performance trajectory and stability throughout the entire CL process, we can compute the Average Incremental R@K for each direction. This metric reflects the area under the performance curve by averaging the mean recall at every incremental step $t$:
\begin{equation}
    \text{AIR}^m = \frac{1}{T} \sum_{t=1}^T \left( \frac{1}{t} \sum_{k=1}^t R^m_{t, k} \right).
\end{equation}

For each query, let $\mathcal{P}(i)$ denote the set of valid ground-truth targets associated with query $i$. In datasets with a single paired annotation, $\mathcal{P}(i)$ contains one element, while datasets such as Flickr30K may contain multiple valid captions per image. All candidates not belonging to $\mathcal{P}(i)$ are treated as negatives. Unless otherwise stated, all metrics are computed independently for image-to-text (I2T) and text-to-image (T2I) retrieval and reported as symmetric averages across both directions.

\paragraph{Recall@K.}
Let $\operatorname{rank}(q_i,r_i)$ denote the one-based rank of the ground-truth target among all candidates sorted by decreasing similarity.
Recall@K is defined as
\[
\operatorname{R@K}
=
\frac{1}{N}
\sum_{i=1}^{N}
\mathbbm{1}\{\operatorname{rank}(q_i,r_i)\le K\}.
\]

\paragraph{Mean Reciprocal Rank.}
Mean Reciprocal Rank (MRR) measures the inverse rank of the first correct retrieval:
\[
\operatorname{MRR}
=
\frac{1}{N}
\sum_{i=1}^{N}
\frac{1}{\operatorname{rank}(q_i,r_i)}.
\]
We report symmetric MRR by averaging the I2T and T2I scores.

\paragraph{Top-$k$ negative margin.}
To assess local separation in the embedding space, we compare the similarity of the positive pair against the hardest nearby negatives.
For a text query $\mathbf{c}_i$, let
\[
s_i^+ = s(\mathbf{c}_i,\mathbf{v}_i)
\]
denote the positive similarity, and let $\mathcal{N}_k(i)$ contain the indices of the $k$ highest-scoring negative images:
\[
\mathcal{N}_k(i)
=
\operatorname{TopK}_{j\neq i}\,
s(\mathbf{c}_i,\mathbf{v}_j).
\]
The Top-$k$ negative margin is
\[
\operatorname{Margin@}k
=
\frac{1}{N}
\sum_{i=1}^{N}
\left(
s_i^+
-
\frac{1}{k}
\sum_{j\in \mathcal{N}_k(i)}
s(\mathbf{c}_i,\mathbf{v}_j)
\right).
\]
We compute the same quantity for I2T retrieval and report the symmetric average across directions. In our experiments we use $k=10$. Larger values indicate better local separation between matched pairs and hard negatives.

\paragraph{Hard-negative violation rate.}
For a query $\mathbf{c}_i$, let
\[
s_i^- = \max_{j\neq i} s(\mathbf{c}_i,\mathbf{v}_j)
\]
denote the hardest-negative similarity.
The violation rate is defined as
\[
\operatorname{Viol}
=
\frac{1}{N}
\sum_{i=1}^{N}
\mathbbm{1}\{s_i^- \ge s_i^+\}.
\]
We compute the metric symmetrically across retrieval directions.
This measures the fraction of queries for which at least one negative candidate scores no lower than the ground-truth match. Lower values indicate stronger local alignment.

\paragraph{I2T/T2I asymmetry gap.}
To quantify directional imbalance between retrieval modes, we compute
\[
\operatorname{Gap@}K
=
\left|
\operatorname{R@K}_{\mathrm{I2T}}
-
\operatorname{R@K}_{\mathrm{T2I}}
\right|.
\]
We use $K=5$, as R@5 is less sensitive than R@1 in datasets where multiple captions or images may be semantically plausible matches.

\paragraph{Top-$k$ neighbor overlap.}
To assess retrieval stability across continual-learning checkpoints, let
$\mathcal{T}_k^{t}(i)$ denote the set of top-$k$ retrieved candidates for query $i$ after learning task $t$, and let
$\mathcal{T}_k^{t-1}(i)$ denote the corresponding set after the previous task.
The neighbor overlap is
\[
\operatorname{Overlap@}k
=
\frac{1}{N}
\sum_{i=1}^{N}
\frac{
|\mathcal{T}_k^{t}(i)\cap \mathcal{T}_k^{t-1}(i)|
}{k}.
\]
Higher overlap indicates greater local neighborhood stability throughout continual adaptation. We use $k=10$.

\paragraph{Linear CKA.}
To evaluate global alignment between image and text representations, we compute linear centered kernel alignment (CKA).
Let
\[
X\in\mathbb{R}^{N\times d_x},
\quad
Y\in\mathbb{R}^{N\times d_y}
\]
denote mean-centered image and text feature matrices.
Linear CKA is defined as
\[
\operatorname{CKA}(X,Y)
=
\frac{
\|X^\top Y\|_F^2
}{
\|X^\top X\|_F \,
\|Y^\top Y\|_F
}.
\]
Higher CKA values indicate stronger global alignment between the image and text embedding spaces.

\clearpage

\section{Additional quantitative results for the main experimental suite}
\label{app:additional_experiments}

\begin{table}[H]
  \centering
  \caption{Cross-modal retrieval performance on our proposed evaluation framework measured by \textbf{Recall@5} at the end of continual training for CLIP ViT-B/16.}
  \label{tab:main_table_r5}
  \resizebox{\linewidth}{!}{
  \small
  \setlength{\tabcolsep}{4pt}
  \begin{tabular}{l cccccccc cccccccc}
  \toprule
  & \multicolumn{8}{c}{\textbf{Text $\rightarrow$ Image}} & \multicolumn{8}{c}{\textbf{Image $\rightarrow$ Text}} \\
  \cmidrule(lr){2-9} \cmidrule(lr){10-17}
  \textbf{Method} 
  & \textbf{Flickr} 
  & \textbf{Lexica} 
  & \textbf{WikiArt} 
  & \textbf{KreaM} 
  & \textbf{Flints} 
  & \textbf{Sketch} 
  & \textbf{ROCOv2} 
  & \textbf{Avg.} 
  & \textbf{Flickr} 
  & \textbf{Lexica} 
  & \textbf{WikiArt} 
  & \textbf{KreaM} 
  & \textbf{Flints} 
  & \textbf{Sketch} 
  & \textbf{ROCOv2} 
  & \textbf{Avg.} \\
  \midrule
  
  \textbf{ZS} 
  &  85.7 & 74.6 & 39.9 & 45.2 & 42.0 & 15.9 & 4.5 & 44.0  
  &  96.7 & 73.5 & 40.2 & 40.7 & 27.0 & 13.8 & 4.3 & 42.3 \\
  
  \textbf{FT} 
  &  92.7 & 83.1 & 61.8 & 52.8 & 75.1 & 22.8 & 15.4 & 57.7  
  &  97.3 & 82.9 & 63.2 & 53.9 & 69.6 & 22.2 & 16.4 & 57.9 \\
  
  \textbf{EWC} 
  &  90.6 & 76.6 & 55.0 & 56.3 & 77.6 & 27.6 & 25.2 &  58.4
  &  96.9 & 74.0 & 53.6 & 56.4 & 74.6 & 27.4 & 24.9 & 58.3 \\
  
  \textbf{Mod-X} 
  &  92.4 & 80.7 & 61.0 & 55.1 & 76.7 & 25.2 & 20.5  & 58.8  
  &  97.3 & 80.2 & 61.6 & 55.9 & 73.8 & 25.4 & 20.5  & 59.2 \\
  
  \textbf{C-CLIP} 
  & 92.3 & 82.7 & 58.1 & 49.4 & 66.1 & 21.2 & 8.7 &  54.1
  & 97.9 & 83.2 & 58.0 & 49.5 & 53.8 & 20.1 & 9.0  & 53.1 \\
  
  \textbf{L2P} 
  &  88.2 & 73.4 & 44.7 & 40.6 & 49.4 & 15.8 & 7.4  &  45.6 
  &  97.0 & 73.9 & 40.3 & 34.1 & 35.5 & 13.1 & 6.9  &  43.0 \\

  \textbf{DKR} 
  &  85.6 & 68.2 & 46.4 & 49.9 & 72.5 & 26.1 & 26.9 & 53.7
  &  93.2 & 60.1 & 39.9 & 47.6 & 67.8 & 22.6 & 26.2 & 51.1 \\
  
  \textbf{TA} 
  & 92.4 & 83.2 & 59.1 & 48.6 & 65.5 & 21.1 & 8.7 &  54.1
  &  97.5 & 83.2 & 59.5 & 47.5 & 57.1 & 20.0 & 9.6 & 53.5 \\

  \textbf{DAR} 
  &  95.5 & 91.6 & 76.5 & 78.1 & 87.8 & 39.5 & 28.8  &  \textbf{71.1}
  & 99.5 & 91.0 & 77.4 & 77.3 & 86.2 & 39.7 & 28.8 & \textbf{71.4} \\

  \bottomrule
  \end{tabular}
  }
\end{table}

\begin{table}[H]
  \centering
  \caption{Cross-modal retrieval performance on \textbf{a reverse of our main evaluation framework} measured by Recall@1 at the end of continual training for CLIP ViT-B/16.
  }
  \label{tab:main_table_reverse}
  \resizebox{\linewidth}{!}{
  \small
  \setlength{\tabcolsep}{4pt}
  \begin{tabular}{l cccccccc cccccccc}
  \toprule
  & \multicolumn{8}{c}{\textbf{Text $\rightarrow$ Image}} & \multicolumn{8}{c}{\textbf{Image $\rightarrow$ Text}} \\
  \cmidrule(lr){2-9} \cmidrule(lr){10-17}
  \textbf{Method} 
  & \textbf{ROCOv2} 
  & \textbf{Sketch} 
  & \textbf{Flints} 
  & \textbf{KreaM} 
  & \textbf{WikiArt} 
  & \textbf{Lexica} 
  & \textbf{Flickr} 
  & \textbf{Avg.} 
  & \textbf{ROCOv2} 
  & \textbf{Sketch} 
  & \textbf{Flints} 
  & \textbf{KreaM} 
  & \textbf{WikiArt} 
  & \textbf{Lexica} 
  & \textbf{Flickr} 
  & \textbf{Avg.} \\
  \midrule
  
  \textbf{ZS} 
  &  1.8 & 5.3 & 16.6 & 22.0 & 20.6 & 53.3 & 62.3  &  26.0
  &  1.5 & 4.2 & 11.1 & 20.2 & 20.8 & 53.0 & 82.0  & 27.5 \\
  
  \textbf{FT} 
  &  5.1 & 8.9 & 35.4 & 27.1 & 38.1 & 68.8 & 75.9 & 37.1
  & 5.4 & 8.1 & 31.6 & 27.9 & 38.2 & 68.8 & 90.6 & 38.7 \\
  
  \textbf{EWC} 
  & 7.0 & 10.1 & 40.1 & 32.7 & 41.0 & 68.3 & 79.6 &  39.8 
  & 6.2 & 8.6 & 32.1 & 31.7 & 37.4 & 68.5 & 92.3  &  39.5 \\
  
  \textbf{Mod-X} 
  &  6.0 & 9.5 & 39.0 & 29.3 & 39.8 & 69.6 & 77.9 & 38.7
  & 6.4 & 8.5 & 32.5 & 29.7 & 38.8 & 69.0 & 91.9  & 9.6 \\
  
  \textbf{C-CLIP} 
  &  3.2 & 8.2 & 30.9 & 25.5 & 34.8 & 65.0 & 74.1 & 34.5
  & 3.0 & 6.6 & 24.4 & 26.1 & 33.5 & 67.7 & 88.8  & 35.7 \\
  
  \textbf{L2P} 
  &  2.2 & 5.9 & 22.3 & 20.6 & 26.8 & 55.3 & 70.2 &  29.0
  & 1.7 & 5.0 & 16.9 & 18.6 & 25.6 & 51.5 & 86.4  &  29.4 \\

  \textbf{DKR} 
  & 6.9 & 10.0 & 41.2 & 33.5 & 43.4 & 69.4 & 80.3 &  40.7
  & 6.0 & 8.0 & 32.4 & 31.4 & 39.4 & 69.3 & 93.4 & 40.0 \\
  
  \textbf{TA} 
  & 3.2 & 8.4 & 32.5 & 24.3 & 34.9 & 62.5 & 73.1 &  34.1
  &  3.7 & 7.9 & 27.5 & 24.5 & 35.2 & 65.0 & 88.3 & 36.0 \\

  \textbf{DAR} 
  &  9.7 & 13.6 & 45.1 & 39.8 & 42.7 & 74.5 & 75.1  &  \textbf{42.9}
  &  10.0 & 13.5 & 43.4 & 39.1 & 43.6 & 74.5 & 88.1 & \textbf{44.6} \\
  
  \bottomrule
  \end{tabular}
  }
\end{table}

\clearpage
\section{Full quantitative results for zero-shot retrieval on our main setup}
\label{app:full_zero_shot}

\begin{table}[H]
\centering
\caption{Zero-shot retrieval performance on COCO~\citep{lin2014microsoft} during continual fine-tuning. We report Recall@1 for Image-to-Text (I2T) and Text-to-Image (T2I), together with performance difference $\Delta$.}
\label{tab:zs_retrieval_coco}
\setlength{\tabcolsep}{4pt}
\renewcommand{\arraystretch}{1.1}
\small
\begin{tabular}{llccccccccc}
\toprule
& & \multicolumn{8}{c}{Task ID} & \multirow{2}{*}{ $\Delta$} \\
\cmidrule(lr){3-10}
Direction & Method & 0 & 1 & 2 & 3 & 4 & 5 & 6 & 7 & \\
\midrule

\multirow{14}{*}{I2T}

& Fine-tuning & 52.4 & 59.4 & 59.3 & 59.7 & 60.4 & 60.2 & 60.0 & 61.0 & +8.6 \\
& EWC & 52.4 & 62.7 & 62.8 & 61.7 & 61.5 & 60.7 & 60.1 & 59.9 & +7.5 \\
& Mod-X & 52.4 & 61.4 & 61.9 & 60.5 & 60.9 & 60.2 & 59.6 & 59.0 & +6.6 \\
& C-CLIP & 52.4 & 57.7 & 58.2 & 58.8 & 59.0 & 59.7 & 60.8 & 59.8 & +7.4 \\
& L2P & 52.4 & 52.2 & 53.0 & 53.5 & 55.0 & 52.2 & 53.6 & 53.4 & +1.0 \\

& DKR & 52.4 & 62.2 & 62.3 & 59.4 & 58.6 & 57.8 & 54.9 & 45.4 & -6.9 \\

& TA & 52.4 & 57.4 & 57.7 & 58.5 & 58.9 & 59.6 & 59.4 & 59.7 & +7.3 \\
& DAR & 52.4 & 61.7 & 61.7 & 61.9 & 61.8 & 61.8 & 61.6 & 61.9 & \textbf{+9.5} \\

\midrule

\multirow{14}{*}{T2I}
& Fine-tuning & 33.2 & 41.6 & 41.3 & 41.5 & 41.9 & 41.9 & 41.7 & 41.6 & +8.4 \\
& EWC & 33.2 & 44.3 & 43.8 & 43.7 & 43.4 & 42.3 & 42.5 & 42.4 & +9.2 \\
& Mod-X & 33.2 & 43.3 & 42.9 & 42.3 & 42.6 & 41.5 & 41.8 & 40.9 & +7.7 \\
& C-CLIP & 33.2 & 38.9 & 39.3 & 40.9 & 41.0 & 41.6 & 41.9 & 41.7 & +8.5 \\
& L2P & 33.2 & 37.1 & 36.9 & 37.8 & 37.7 & 37.8 & 37.4 & 37.1 & +3.9 \\

& DKR & 33.1 & 44.2 & 43.1 & 41.7 & 40.7 & 38.4 & 37.2 & 31.4 & -1.7 \\
& TA & 33.2 & 38.7 & 38.9 & 40.6 & 40.5 & 40.6 & 40.8 & 41.0 & +7.8 \\
& DAR  & 33.2 &44.1 & 44.1 & 44.2 & 44.2 & 44.2 & 44.1 & 44.1 & \textbf{+10.2} \\

\bottomrule
\end{tabular}
\end{table}

\begin{table}[H]
\centering
\caption{Zero-shot retrieval performance on NoCaps~\citep{agrawal2019nocaps} during continual fine-tuning. We report Recall@1 for Image-to-Text (I2T) and Text-to-Image (T2I), together with performance difference $\Delta$.}
\label{tab:no_caps}
\setlength{\tabcolsep}{4pt}
\renewcommand{\arraystretch}{1.1}
\small
\begin{tabular}{llccccccccc}
\toprule
& & \multicolumn{8}{c}{Task ID} & \multirow{2}{*}{ $\Delta$} \\
\cmidrule(lr){3-10}
Direction & Method & 0 & 1 & 2 & 3 & 4 & 5 & 6 & 7 & \\
\midrule

\multirow{14}{*}{I2T}

& Fine-tuning & 71.7 & 78.7 & 79.4 & 79.4 & 79.9 & 80.0 & 79.4 & 80.3 & +8.6 \\
& EWC & 71.7 & 82.0 & 81.5 & 80.8 & 80.2 & 79.6 & 78.3 & 76.1 & +4.4 \\
& Mod-X & 71.7 & 81.1 & 80.9 & 80.4 & 81.0 & 80.2 & 80.0 & 79.8 & +8.1 \\
& C-CLIP  & 71.7 & 76.1 & 77.2 & 78.3 & 79.1 & 79.9 & 80.0 & 80.0 & +8.3 \\
& L2P & 71.7 & 72.0 & 72.7 & 73.8 & 74.9 & 71.9 & 73.2 & 73.6 & +0.9 \\

& DKR & 71.7 & 81.6 & 81.8 & 80.7 & 80.0 & 78.9 & 75.2 & 68.2 & -3.5 \\

& TA  & 71.7 & 76.2 & 76.6 & 78.1 & 78.4 & 79.0 & 78.9 & 79.7& +8.0 \\

& DAR & 71.7 & 81.6 & 81.8 & 81.9 & 81.9 & 81.8 & 81.4 & 81.4 & \textbf{+9.7} \\

\midrule

\multirow{14}{*}{T2I}
& Fine-tuning & 46.7 & 56.1 & 55.9 & 56.8 & 57.4 & 57.4 & 56.8 & 56.6 & +9.9 \\
& EWC & 46.7 & 59.9 & 58.8 & 57.8 & 57.3 & 56.1 & 55.6 & 52.3 & +5.6 \\
& Mod-X & 46.7 & 58.2 & 57.8 & 57.8 & 58.2 & 57.4 & 57.1 & 55.9 & +9.2 \\ 
& C-CLIP& 46.7 & 53.1 & 53.6 & 56.0 & 56.0 & 56.6 & 56.7 & 56.3 & +9.6 \\ 
& L2P & 46.7 & 51.3 & 51.1 & 52.5 & 52.4 & 53.0 & 52.2 & 51.3 & +4.6 \\ 

& DKR & 46.7 & 59.7 & 58.8 & 57.9 & 57.0 & 55.0 & 52.2 & 45.4 & -1.3 \\

& TA& 46.7 & 53.0 & 53.2 & 55.1 & 55.3 & 55.3 & 55.3 & 55.6 & +8.9 \\

& DAR & 46.7 &60.4 & 60.4 & 60.7 & 60.6 & 60.6 & 60.2 & 60.2 & \textbf{+13.5} \\

\bottomrule
\end{tabular}
\end{table}

\section{Full quantitative results for zero-shot classification}
\label{app:full_classification_results}

\begin{table}[H]
\centering
\caption{Zero-shot accuracy on ImageNet during continual fine-tuning.}
\label{tab:zs_accuracy_imagenet}
\setlength{\tabcolsep}{4pt}
\renewcommand{\arraystretch}{1.1}
\small
\begin{tabular}{lccccccccc}
\toprule
Method & 0 & 1 & 2 & 3 & 4 & 5 & 6 & 7 & $\Delta$ \\
\midrule
Fine-tuning & 68.1 & 68.8 & 68.8 & 67.4 & 67.8 & 65.9 & 65.6 & 66.5 & -1.6 \\
EWC & 68.1 & 67.9 & 67.6 & 66.6 & 66.1 & 63.7 & 62.5 & 63.1 & -5.0 \\
Mod-X & 68.1 & 68.6 & 68.3 & 66.2 & 66.4 & 63.2 & 62.7 & 63.2 & -4.9 \\
C-CLIP & 68.1 & 69.2 & 69.1 & 68.9 & 68.7 & 67.9 & 67.7 & 67.6 & \textbf{-0.5} \\
L2P & 68.1 & 65.2 & 66.5 & 66.5 & 67.4 & 64.8 & 65.9 & 66.3 & -1.8 \\

DKR & 68.1 & 68.0 & 66.5 & 63.7 & 62.2 & 56.6 & 53.9 & 48.2 & -19.9 \\

TA & 68.1 & 69.0 & 68.9 & 68.5 & 68.5 & 67.6 & 67.3 & 67.0 & -1.1 \\
DAR  & 68.1 & 64.3 & 64.2 & 64.2 & 64.2 & 64.2 & 64.2 & 64.2 & -3.9 \\

\bottomrule
\end{tabular}
\end{table}

\begin{table}[H]
\centering
\caption{Zero-shot accuracy on CIFAR100 during continual fine-tuning.}
\label{tab:zs_accuracy_cifar100}
\setlength{\tabcolsep}{4pt}
\renewcommand{\arraystretch}{1.1}
\small
\begin{tabular}{lccccccccc}
\toprule
Method & 0 & 1 & 2 & 3 & 4 & 5 & 6 & 7 & $\Delta$ \\
\midrule
Fine-tuning & 68.4 & 68.7 & 68.4 & 69.7 & 69.9 & 68.8 & 69.6 & 71.4 & +3 \\
EWC & 68.4 & 68.2 & 68.5 & 69.2 & 67.7 & 66.9 & 65.4 & 68.3 & -0.1 \\

Mod-X & 68.4 & 69.0 & 68.8 & 69.5 & 69.1 & 67.0 & 67.8 & 67.7 & -0.7 \\
C-CLIP & 68.4 & 69.7 & 69.7 & 70.7 & 70.6 & 71.1 & 72.0 & 72.7 & \textbf{+4.3} \\
L2P & 68.4 & 65.9 & 66.8 & 66.6 & 68.0 & 63.5 & 66.8 & 64.4 & -4.0 \\

DKR & 68.4 & 68.4 & 67.3 & 67.5 & 65.4 & 60.2 & 59.7 & 45.2 & -23.2 \\

TA & 68.4 & 68.7 & 68.6 & 69.3 & 69.4 & 69.4 & 69.7 & 70.1 & +1.7 \\
DAR  & 68.4 & 65.3 & 65.4 & 65.4 & 65.4 & 65.4 & 65.9 & 65.8 & 0.0 \\

\bottomrule
\end{tabular}
\end{table}

\begin{table}[H]
\centering
\caption{Zero-shot accuracy on EuroSAT during continual fine-tuning.}
\label{tab:zs_accuracy_eurosat}
\setlength{\tabcolsep}{4pt}
\renewcommand{\arraystretch}{1.1}
\small
\begin{tabular}{lccccccccc}
\toprule
Method & 0 & 1 & 2 & 3 & 4 & 5 & 6 & 7 & $\Delta$ \\
\midrule
Fine-tuning & 54.0 & 54.3 & 56.9 & 52.6 & 56.5 & 50.6 & 51.1 & 55.8 & +1.8 \\
EWC & 54.0 & 51.4 & 56.1 & 53.1 & 56.2 & 47.4 & 47.7 & 55.0 & +1.0 \\
Mod-X & 54.0 & 51.5 & 56.3 & 49.7 & 54.9 & 43.9 & 46.8 & 50.4 & -3.6 \\

C-CLIP & 54.0 & 53.5 & 56.4 & 56.1 & 59.2 & 57.1 & 58.0 & 56.9 & \textbf{+2.9} \\
L2P & 54.0 & 48.3 & 51.4 & 52.1 & 59.3 & 53.7 & 52.5 & 54.9 & +0.1 \\
DKR & 54.0 & 50.9 & 54.9 & 50.2 & 47.3 & 33.4 & 33.8 & 32.8 & -21.2 \\
TA & 54.0 & 54.1 & 56.4 & 55.4 & 55.6 & 52.1 & 53.8 & 53.6 & -0.4 \\
DAR  & 54.0 & 48.9 & 49.0 & 49.0 & 49.0 & 49.0 & 49.0 & 49.0 & -5.0 \\

\bottomrule
\end{tabular}
\end{table}

\begin{table}[H]
\centering
\caption{Zero-shot accuracy on DomainNet during continual fine-tuning.}
\label{tab:zs_accuracy_domainnet}
\setlength{\tabcolsep}{4pt}
\renewcommand{\arraystretch}{1.1}
\small
\begin{tabular}{lccccccccc}
\toprule
Method & 0 & 1 & 2 & 3 & 4 & 5 & 6 & 7 &  $\Delta$ \\
\midrule
Fine-tuning & 56.8 & 56.8 & 56.8 & 56.1 & 56.1 & 56.1 & 56.3 & 55.6 & -1.2 \\
EWC & 56.8 & 56.1 & 55.6 & 55.5 & 55.0 & 55.0 & 55.0 & 54.3 & -2.5 \\
Mod-X & 56.8 & 56.6 & 56.2 & 55.5 & 55.3 & 55.3 & 55.4 & 55.0 & -1.8 \\

C-CLIP & 56.8 & 56.9 & 56.8 & 56.6 & 56.6 & 56.7 & 56.9 & 56.7 & \textbf{-0.1} \\

L2P & 56.8 & 56.1 & 56.5 & 56.3 & 56.8 & 55.7 & 56.4 & 54.6 & -2.2 \\
DKR & 56.8 & 56.3 & 55.4 & 54.5 & 53.1 & 52.0 & 50.6 & 45.9 & -10.9 \\
TA & 56.8 & 56.9 & 57.0 & 56.7 & 56.7 & 56.5 & 56.6 & 56.2 & -0.6 \\
DAR & 56.8 & 54.6 & 54.6 & 54.6 & 54.7 & 54.7 & 54.9 & 54.9 & -1.9 \\

\bottomrule
\end{tabular}
\end{table}

\section{Full quantitative results alternative tasks sequences}
\label{app:sec_full_results_long_sequences}

\begin{table}[H]
\centering
\caption{
Recall@1 after continual training on a longer 14-task sequence using CLIP ViT-B/16. 
Each original dataset is split into two consecutive subtasks, denoted by suffixes 0 and 1, and the sequence groups both halves of each domain together 
(e.g., Flickr0, Flickr1, Lexica0, Lexica1, ...). 
Results are reported for both Text-to-Image and Image-to-Text retrieval, with averages computed across all subtasks. 
This setting evaluates whether methods remain robust when each domain is revisited immediately through a second split.
}
\label{tab:main_table_halves_blocked}
\small
\setlength{\tabcolsep}{3pt}
\resizebox{\textwidth}{!}{%
\begin{tabular}{lccccccccccccccc}
\toprule
\multicolumn{16}{c}{\textbf{Text $\rightarrow$ Image}} \\
\midrule
\textbf{Method}
& \textbf{Flickr0} & \textbf{Flickr1}
& \textbf{Lexica0} & \textbf{Lexica1}
& \textbf{WikiArt0} & \textbf{WikiArt1}
& \textbf{KreaM0} & \textbf{KreaM1}
& \textbf{Flints0} & \textbf{Flints1}
& \textbf{Sketch0} & \textbf{Sketch1}
& \textbf{ROCO0} & \textbf{ROCO1}
& \textbf{Avg.} \\
\midrule
ZS     & 62.8 & 62.3 & 60.1 & 60.0 & 26.4 & 26.9 & 29.2 & 28.3 & 25.4 & 24.0 & 7.6 & 8.8 & 2.5 & 2.5 & 30.3 \\
FT     & 73.4 & 74.3 & 69.9 & 71.4 &	44.6 &	44.9 & 35.5 & 34.0 &	51.3 & 50.6 & 11.3	& 14.6	& 9.4 &	9.02 & 42.4 \\

EWC    & 67.6 & 69.6 & 59.7 & 60.0 & 36.0 & 36.2 & 34.9 & 34.7 & 51.3 & 50.7 & 14.7 & 14.1 & 15.3 & 15.3 & 42.2 \\
Mod-X  & 72.8 & 74.2 & 66.1 & 69.4 & 43.6 & 44.1 & 36.5 & 35.4 & 53.0 & 53.8 & 12.0 & 14.6 & 12.7 & 12.3 & 43.2 \\

C-CLIP & 73.4 & 74.4 & 67.8 & 69.6 & 42.5 & 43.1 & 32.9 & 32.0 & 46.3 & 45.5 & 10.5 & 12.8 & 6.2 & 6.4 & 41.0 \\
L2P    & 65.8 & 67.6 & 61.1 & 61.6 & 31.8 & 32.7 & 27.3 & 26.8 & 29.7 & 27.4 & 7.5 & 10.7 & 4.5 & 4.1 & 32.8 \\
DKR & 57.7 & 59.2 & 46.6 & 46.3 & 24.5 & 25.4 & 28.8 & 28.5 & 44.5 & 46.1 & 12.8 & 12.2 & 15.7 & 15.4 & 32.7 \\
TA     &70.0 & 70.0 & 65.7 & 66.6 & 39.4 & 38.8 & 27.4 & 26.8 & 39.3 & 39.0 & 9.4 & 12.2 & 3.9 & 4.0 & 36.6 \\
\textbf{DAR}
       & 79.9 & 81.7 & 83.1 & 81.5 & 58.3 & 58.3 & 54.0 & 53.9 & 63.1 & 63.1 & 21.1 & 21.1 & 16.6 & 16.8 & \textbf{55.2} \\
\midrule

\multicolumn{16}{c}{\textbf{Image $\rightarrow$ Text}} \\
\midrule
\textbf{Method}
& \textbf{Flickr0} & \textbf{Flickr1}
& \textbf{Lexica0} & \textbf{Lexica1}
& \textbf{WikiArt0} & \textbf{WikiArt1}
& \textbf{KreaM0} & \textbf{KreaM1}
& \textbf{Flints0} & \textbf{Flints1}
& \textbf{Sketch0} & \textbf{Sketch1}
& \textbf{ROCO0} & \textbf{ROCO1}
& \textbf{Avg.} \\
\midrule
ZS     & 72.9 & 74.9 & 59.2 & 59.6 & 26.7 & 26.8 & 27.0 & 25.8 & 14.6 & 16.2 & 6.8 & 6.8 & 2.3 & 2.2 & 30.1 \\

FT & 81.7 & 81.1 & 71.0 & 70.0 & 46.2 & 46.7 & 36.5 & 35.9 &	47.7 & 46.2 & 12.9 & 12.1 & 9.7 & 9.9 & 43.4 \\
EWC    & 75.6 & 77.2 & 56.9 & 56.3 & 32.7 & 34.3 & 34.7 & 35.8 & 47.1 & 48.2 & 12.6 & 12.7 & 14.7 & 15.2 & 39.2 \\
Mod-X  & 80.3 & 80.8 & 67.1 & 67.5 & 44.0 & 44.4 & 37.1 & 37.0 & 49.1 & 49.7 & 13.7 & 12.7 & 12.7 & 12.5 & 43.8 \\

C-CLIP &80.7 & 80.8 & 70.1 & 69.9 & 43.3 & 43.4 & 33.7 & 32.9 & 40.3 & 39.7 & 11.2 & 11.2 & 6.5 & 6.9 & 40.8 \\
L2P    &75.7 & 77.8 & 60.1 & 59.1 & 29.0 & 29.3 & 23.9 & 23.5 & 22.5 & 23.3 & 7.3 & 7.6 & 3.6 & 3.6 & 32.5 \\
DKR & 64.9 & 65.3 & 39.6 & 39.3 & 20.0 & 20.8 & 25.8 & 26.1 & 38.0 & 39.1 & 10.9 & 10.0 & 14.6 & 14.5 & 29.9 \\
Meging-TA     & 78.7 & 79.6 & 67.5 & 64.5 & 38.6 & 39.5 & 27.7 & 27.5 & 37.7 & 36.1 & 10.8 & 12.0 & 4.8 & 5.2 & 38.6 \\
\textbf{DAR}
       & 87.2 & 88.4 & 81.7 & 81.4 & 59.6 & 58.6 & 53.4 & 54.2 & 61.1 & 62.3 & 20.8 & 20.3 & 16.0 & 16.4&  \textbf{55.8} \\

\bottomrule
\end{tabular}%
}
\end{table}

\begin{table}[H]
\centering
\caption{
Recall@1 after continual training on a longer 14-task interleaved sequence using CLIP ViT-B/16.  Each original dataset is split into two subtasks, denoted by suffixes 0 and 1, but all first splits are seen before the corresponding second splits 
(Flickr0, Lexica0, ..., ROCO0, followed by Flickr1, Lexica1, ..., ROCO1). 
Results are reported for both Text-to-Image and Image-to-Text retrieval, with averages computed across all subtasks. This setting tests robustness to delayed domain revisitation and evaluates whether methods retain knowledge across a longer gap before seeing the second split of each domain.
}
\label{tab:main_table_halves_interleaved}
\small
\setlength{\tabcolsep}{3pt}
\resizebox{\textwidth}{!}{%
\begin{tabular}{lccccccccccccccc}
\toprule
\multicolumn{16}{c}{\textbf{Text $\rightarrow$ Image}} \\
\midrule
\textbf{Method}
& \textbf{Flickr0} 
& \textbf{Lexica0} 
& \textbf{WikiArt0} 
& \textbf{KreaM0} 
& \textbf{Flints0} 
& \textbf{Sketch0} 
& \textbf{ROCO0}
& \textbf{Flickr1} 
& \textbf{Lexica1} 
& \textbf{WikiArt1} 
& \textbf{KreaM1} 
& \textbf{Flints1} 
& \textbf{Sketch1} 
& \textbf{ROCO1}
& \textbf{Avg.} \\
\midrule
ZS     & 62.8 & 60.1 & 26.4 & 29.2 & 25.4 & 7.6 & 2.5 & 62.3 & 60.0 & 26.9 & 28.3 & 24.0 & 8.8 & 2.5 & 30.5 \\
FT     & 74.7 & 72.6 & 46.7 & 36.9 & 52.3 & 11.6 & 9.1 & 75.6 & 74.3 & 47.1 & 35.4 & 51.2 & 14.3 & 8.8 & 43.6 \\
EWC    & 73.9 & 68.7 & 46.8 & 41.9 & 57.5 & 17.2 & 15.2 & 76.5 & 70.8 & 47.6 & 40.8 & 57.7 & 17.8 & 15.4 & 46.6 \\
Mod-X  & 75.5 & 73.2 & 48.6 & 39.6 & 56.2 & 13.9 & 12.4 & 76.6 & 75.5 & 49.4 & 38.8 & 54.5 & 16.0 & 12.2 & 45.5 \\
C-CLIP & 74.7 & 70.6 & 43.6 & 32.8 & 46.9 & 11.7 & 5.9 & 75.2 & 71.4 & 44.2 & 32.1 & 46.1 & 13.0 & 6.1 & 41.7 \\
L2P    & 67.6 & 62.3 & 31.6 & 26.9 & 32.1 & 7.6 & 4.0 & 68.9 & 62.9 & 32.7 & 25.9 & 30.1 & 9.9 & 4.0 & 33.9 \\
DKR & 69.6 & 62.5 & 41.8 & 38.8 & 54.0 & 17.3 & 15.4 & 70.5 & 64.5 & 42.7 & 38.7 & 54.5 & 17.2 & 16.4 & 43.1\\
TA     & 71.8 & 62.8 & 39.4 & 27.6 & 39.3 & 10.0 & 3.6 & 71.7 & 62.3 & 38.5 & 26.4 & 38.9 & 12.2 & 3.8 & 36.9 \\
\textbf{DAR}
       & 79.5 & 82.4 & 55.2 & 48.7 & 61.8 & 18.9 & 14.1 & 79.9 & 81.4 & 55.7 & 48.7 & 60.5 & 20.3 & 14.2 & \textbf{51.4} \\
\midrule
\multicolumn{16}{c}{\textbf{Image $\rightarrow$ Text}} \\
\midrule
\textbf{Method}
& \textbf{Flickr0} 
& \textbf{Lexica0} 
& \textbf{WikiArt0} 
& \textbf{KreaM0} 
& \textbf{Flints0} 
& \textbf{Sketch0} 
& \textbf{ROCO0}
& \textbf{Flickr1} 
& \textbf{Lexica1} 
& \textbf{WikiArt1} 
& \textbf{KreaM1} 
& \textbf{Flints1} 
& \textbf{Sketch1} 
& \textbf{ROCO1}
& \textbf{Avg.} \\
\midrule
ZS     & 72.9 & 59.2 & 26.6 & 27.0 & 14.6 & 6.1 & 2.3 & 74.9 & 59.6 & 26.8 & 25.8 & 16.2 & 6.8 & 2.2 & 30.1 \\

FT     & 81.7 & 73.6 & 49.2 & 37.1 & 48.5 & 12.9 & 9.3 & 82.9 & 72.5 & 49.4 & 36.5 & 48.5 & 12.3 & 9.8 & 44.6 \\
EWC    & 82.3 & 67.7 & 46.4 & 41.5 & 55.3 & 17.5 & 14.8 & 82.6 & 68.7 & 47.0 & 42.2 & 54.8 & 16.9 & 15.2 & 49.2 \\
Mod-X  & 82.8 & 73.0 & 50.4 & 40.0 & 51.0 & 15.1 & 12.3 & 83.7 & 73.1 & 50.2 & 39.8 & 54.9 & 14.5 & 12.6 & 46.7 \\
C-CLIP & 82.2 & 71.5 & 44.2 & 34.1 & 41.0 & 10.6 & 6.1 & 82.1 & 72.0 & 44.5 & 33.8 & 39.1 & 10.5 & 6.5 & 41.6 \\
L2P    & 76.3 & 61.2 & 29.6 & 23.0 & 22.6 & 6.9 & 3.2 & 79.6 & 60.5 & 30.2 & 22.5 & 23.1 & 7.6 & 3.2 & 32.8 \\
DKR & 75.8 & 57.1 & 37.2 & 36.7 & 49.5 & 17.0 & 14.5 & 78.8 & 58.8 & 38.8 & 36.9 & 52.1 & 16.6 & 14.6 & 41.7 \\
TA     & 79.3 & 66.2 & 38.6 & 27.1 & 36.4 & 10.7 & 4.6 & 80.0 & 63.4 & 38.9 & 27.0 & 35.3 & 11.6 & 5.1 & 37.6 \\
\textbf{DAR} & 86.1 & 82.2 & 57.3 & 49.1 & 60.2 & 18.9 & 13.8 & 87.7 & 80.6 & 57.4 & 49.3 & 60.4 & 19.4 & 13.7 & \textbf{52.6} \\
       
\bottomrule
\end{tabular}%
}
\end{table}

\section{Summary of related work evaluation protocols}
\label{app:summary_of_related_protocols}

\begin{table}[H]
\centering
\caption{Comparison of Continual Retrieval Benchmarks, evaluated each suite against our proposed desiderata: 
\textbf{Div.} (Domain Diversity),  
\textbf{Spec.} (Semantic Specificity),
\textbf{Curr.} (Calibrated Curriculum),
\textbf{Rob.} (OOD Robustness)
\textbf{Acc.} (Accessibility).
}
\label{tab:app_benchmark_comparision}
\small
\setlength{\tabcolsep}{3pt}
\renewcommand{\arraystretch}{1.8} 
\begin{tabularx}{\textwidth}{p{2.4cm} @{\extracolsep{\fill}} p{2.2cm} p{2.2cm} p{2.2cm} p{2.2cm} p{2.1cm}}
\toprule
\textbf{Benchmark} & \textbf{Div.} & \textbf{Spec.} & \textbf{Curr.} & \textbf{Rob.} & \textbf{Acc.} \\
\midrule

\addlinespace

Wang et al. \newline \citep{kai2021cross_modal_retrieval} 
& \xmarkr \newline \scriptsize Natural images only (COCO/VG)
& \cmarkg \newline \scriptsize High-information target pairs
& \xmarkr \newline \scriptsize Only 3 sequential tasks
& \xmarkr \newline \scriptsize No zero-shot/OOD testing
& \cmarkg \newline \scriptsize Small-scale and accessible \\

\addlinespace

Ni et al. \newline \citep{ni2023off_diagonal} 
& \xmarkr \newline \scriptsize Limited to Natural/E-commerce
& \xmarkr \newline \scriptsize High overlap in Flickr/COCO
& \xmarkr \newline \scriptsize Streaming focus; no curriculum
& \xmarkr \newline \scriptsize No cross-domain stability eval
& \cmarkg \newline \scriptsize Lightweight fine-tuning \\ 

\addlinespace

Cui et al. \newline \citep{cui2024knowledge_rectification} 
& \cmarkg \newline \scriptsize Diverse (Natural, e-commerce, etc.)
& \cmarkg \newline \scriptsize Precise semantic targets
& \xmarkr \newline \scriptsize No structured task ordering
& \xmarkr \newline \scriptsize Lacks OOD classification
& \cmarkg \newline \scriptsize Publicly available \\

\addlinespace

Garg et al. \newline \citep{garg2024tic_clip} 
& \cmarkg \newline \scriptsize Diverse although the domain boundaries are not clearly defined. 
& \xmarkr \newline \scriptsize Significant web-scale noise
& \cmarkg \newline \scriptsize Temporal-based ordering
& \cmarkg \newline \scriptsize Extensive OOD suite
& \xmarkr \newline \scriptsize Prohibitive compute costs \\

\addlinespace

Liu et al. \newline \citep{c_clip2025} 
& \cmarkg \newline \scriptsize Diverse (Flickr, WikiArt, etc.)
& \xmarkr \newline \scriptsize Redundant pairs (e.g., Oxford Pets)
& \xmarkr \newline \scriptsize Arbitrary/Random task sequence
& \cmarkg \newline \scriptsize HaVG \& Cls.
& \xmarkr \newline \scriptsize Computationally tractable but no code \\

\addlinespace

\midrule

\textbf{Our Framework} 
& \cmarkg \newline \scriptsize 7 heterogeneous domains (Med, Art, etc.)
& \cmarkg \newline \scriptsize Curated high-density captions
& \cmarkg \newline \scriptsize ZS-calibrated difficulty order
& \cmarkg \newline \scriptsize Multi-dataset Ret. \& Cls.
& \cmarkg \newline \scriptsize Moderate scale; Open-source \\

\addlinespace

\bottomrule
\end{tabularx}
\end{table}

\begin{table}[H]
\centering
\caption{Summary of training and evaluation datasets across selected vision-language CL papers. }
\label{tab:vl_cl_datasets}
\small
\setlength{\tabcolsep}{4pt}

\renewcommand{\arraystretch}{1}
\begin{tabularx}{\textwidth}{p{2.5cm} X X X X X}
\toprule
\textbf{Paper} 
& \textbf{Training Data} 
& \textbf{Training Domains Approximation \newline (no. unique vs total)}
& \textbf{Eval ID} 
& \textbf{Eval OOD Classification} 
& \textbf{Eval OOD Retrieval} \\
\midrule

\textbf{Our framework} 
& \textbf{Retrieval:} Flickr30K, Lexica-SD, WikiArt, Kream, Flintstones, Sketch, ROCOv2 
& 7/7: Natural images, synthetic, art, fashion, cartoons, sketches, medical 
& Same datasets 
& ImageNet, CIFAR100, EuroSAT, DomainNet 
& COCO ~\citep{lin2014microsoft}, NoCaps~\citep{agrawal2019nocaps} \\

C-CLIP~\citep{c_clip2025}
& \textbf{Retrieval:} Flickr30K, COCO, Pets, Lexica, Simpsons, WikiArt, Kream, Sketch 
& 6/8: Natural images, synthetic, cartoons, art, e-commerce, sketches 
& Same datasets 
& CIFAR100, ImageNet, Flowers, DTD, Food101, Stanford Cars 
& HAVG \\

DKR~\citep{cui2024knowledge_rectification} 
& \textbf{Retrieval:} MS-COCO, Flickr30K, IAPR TC-12, EC, RSICD 
& 4/5: Natural images, remote sensing, e-commerce/products, general web images 
& Same datasets 
& -- 
& Train on EC splits $\rightarrow$ test on unseen MS-COCO, Flickr30K \\

TIC-CLIP~\citep{garg2024tic_clip}
& \textbf{Retrieval:} TIC-DataComp, TIC-YFCC, TIC-RedCaps 
& 2/3: Large-scale web data, diverse real-world domains 
& Classification: TIC-DataComp-Net; Retrieval: TIC-DataComp-Retrieval, TIC-YFCC Retrieval and TIC-RedCaps 
& Over 28 datasets: ImageNet, Food101, MNIST, Oxford-Flowers, Stanford Cars, SUN-97, Oxford-Pet, ObjectNet, and more.
& Flickr30k \\

Cross-modal Retrieval~\citep{kai2021cross_modal_retrieval} 
& \textbf{Retrieval:} Sequential Visual Genome (SeViGe), Sequential MS-COCO (SeCOCO) 
& 1/2: Natural images
& Same datasets 
& -- 
& --\\

Mod-X~\citep{ni2023off_diagonal} 
& \textbf{Retrieval:} COCO (initial training) + Flickr30K (streaming); also ECommerce-T2I in extended experiments 
& 2/3: Natural images, e-commerce/products 
& Same datasets
& -- 
& Train on COCO/Flickr $\rightarrow$ test on unseen ECommerce-T2I \\

\bottomrule
\end{tabularx}
\end{table}

\newpage
\section*{NeurIPS Paper Checklist}

\begin{enumerate}

\item {\bf Claims}
    \item[] Question: Do the main claims made in the abstract and introduction accurately reflect the paper's contributions and scope?
    \item[] Answer: \answerYes{} 
    \item[] Justification: The abstract and introduction state the paper's main contributions: a continual multimodal retrieval evaluation framework, a systematic empirical comparison of CL methods, and Dynamic Adapter Routing (DAR). These claims are supported by the main results in Section \ref{sec:experiments} and \Cref{tab:main_table}.
    \item[] Guidelines:
    \begin{itemize}
        \item The answer \answerNA{} means that the abstract and introduction do not include the claims made in the paper.
        \item The abstract and/or introduction should clearly state the claims made, including the contributions made in the paper and important assumptions and limitations. A \answerNo{} or \answerNA{} answer to this question will not be perceived well by the reviewers. 
        \item The claims made should match theoretical and experimental results, and reflect how much the results can be expected to generalize to other settings. 
        \item It is fine to include aspirational goals as motivation as long as it is clear that these goals are not attained by the paper. 
    \end{itemize}

\item {\bf Limitations}
    \item[] Question: Does the paper discuss the limitations of the work performed by the authors?
    \item[] Answer: \answerYes{}
    \item[] Justification: 
    We include the limitation discussion in \Cref{sec:conclusions}.
    
    \item[] Guidelines:
    \begin{itemize}
        \item The answer \answerNA{} means that the paper has no limitation while the answer \answerNo{} means that the paper has limitations, but those are not discussed in the paper. 
        \item The authors are encouraged to create a separate ``Limitations'' section in their paper.
        \item The paper should point out any strong assumptions and how robust the results are to violations of these assumptions (e.g., independence assumptions, noiseless settings, model well-specification, asymptotic approximations only holding locally). The authors should reflect on how these assumptions might be violated in practice and what the implications would be.
        \item The authors should reflect on the scope of the claims made, e.g., if the approach was only tested on a few datasets or with a few runs. In general, empirical results often depend on implicit assumptions, which should be articulated.
        \item The authors should reflect on the factors that influence the performance of the approach. For example, a facial recognition algorithm may perform poorly when image resolution is low or images are taken in low lighting. Or a speech-to-text system might not be used reliably to provide closed captions for online lectures because it fails to handle technical jargon.
        \item The authors should discuss the computational efficiency of the proposed algorithms and how they scale with dataset size.
        \item If applicable, the authors should discuss possible limitations of their approach to address problems of privacy and fairness.
        \item While the authors might fear that complete honesty about limitations might be used by reviewers as grounds for rejection, a worse outcome might be that reviewers discover limitations that aren't acknowledged in the paper. The authors should use their best judgment and recognize that individual actions in favor of transparency play an important role in developing norms that preserve the integrity of the community. Reviewers will be specifically instructed to not penalize honesty concerning limitations.
    \end{itemize}

\item {\bf Theory assumptions and proofs}
    \item[] Question: For each theoretical result, does the paper provide the full set of assumptions and a complete (and correct) proof?
    \item[] Answer: \answerNA{}
    \item[] Justification: The paper does not present theoretical results, theorems, or formal proof claims.

    \item[] Guidelines:
    \begin{itemize}
        \item The answer \answerNA{} means that the paper does not include theoretical results. 
        \item All the theorems, formulas, and proofs in the paper should be numbered and cross-referenced.
        \item All assumptions should be clearly stated or referenced in the statement of any theorems.
        \item The proofs can either appear in the main paper or the supplemental material, but if they appear in the supplemental material, the authors are encouraged to provide a short proof sketch to provide intuition. 
        \item Inversely, any informal proof provided in the core of the paper should be complemented by formal proofs provided in appendix or supplemental material.
        \item Theorems and Lemmas that the proof relies upon should be properly referenced. 
    \end{itemize}

    \item {\bf Experimental result reproducibility}
    \item[] Question: Does the paper fully disclose all the information needed to reproduce the main experimental results of the paper to the extent that it affects the main claims and/or conclusions of the paper (regardless of whether the code and data are provided or not)?
    \item[] Answer: \answerYes{}
    \item[] Justification: The paper specifies the benchmark construction, task sequence, datasets, evaluation metrics, baselines, model backbone, optimization protocol, and DAR hyperparameters in Sections \ref{sec:framework}, \ref{sec:experiments}, and Appendix \ref{app:implementation_details}.

    \item[] Guidelines:
    \begin{itemize}
        \item The answer \answerNA{} means that the paper does not include experiments.
        \item If the paper includes experiments, a \answerNo{} answer to this question will not be perceived well by the reviewers: Making the paper reproducible is important, regardless of whether the code and data are provided or not.
        \item If the contribution is a dataset and\slash or model, the authors should describe the steps taken to make their results reproducible or verifiable. 
        \item Depending on the contribution, reproducibility can be accomplished in various ways. For example, if the contribution is a novel architecture, describing the architecture fully might suffice, or if the contribution is a specific model and empirical evaluation, it may be necessary to either make it possible for others to replicate the model with the same dataset, or provide access to the model. In general. releasing code and data is often one good way to accomplish this, but reproducibility can also be provided via detailed instructions for how to replicate the results, access to a hosted model (e.g., in the case of a large language model), releasing of a model checkpoint, or other means that are appropriate to the research performed.
        \item While NeurIPS does not require releasing code, the conference does require all submissions to provide some reasonable avenue for reproducibility, which may depend on the nature of the contribution. For example
        \begin{enumerate}
            \item If the contribution is primarily a new algorithm, the paper should make it clear how to reproduce that algorithm.
            \item If the contribution is primarily a new model architecture, the paper should describe the architecture clearly and fully.
            \item If the contribution is a new model (e.g., a large language model), then there should either be a way to access this model for reproducing the results or a way to reproduce the model (e.g., with an open-source dataset or instructions for how to construct the dataset).
            \item We recognize that reproducibility may be tricky in some cases, in which case authors are welcome to describe the particular way they provide for reproducibility. In the case of closed-source models, it may be that access to the model is limited in some way (e.g., to registered users), but it should be possible for other researchers to have some path to reproducing or verifying the results.
        \end{enumerate}
    \end{itemize}

\item {\bf Open access to data and code}
    \item[] Question: Does the paper provide open access to the data and code, with sufficient instructions to faithfully reproduce the main experimental results, as described in supplemental material?
    \item[] Answer: \answerYes{}
    \item[] Justification: The paper provides an anonymized repository for the evaluation framework and implementation, and the benchmark is constructed from publicly accessible datasets.

    \item[] Guidelines:
    \begin{itemize}
        \item The answer \answerNA{} means that paper does not include experiments requiring code.
        \item Please see the NeurIPS code and data submission guidelines (\url{https://neurips.cc/public/guides/CodeSubmissionPolicy}) for more details.
        \item While we encourage the release of code and data, we understand that this might not be possible, so \answerNo{} is an acceptable answer. Papers cannot be rejected simply for not including code, unless this is central to the contribution (e.g., for a new open-source benchmark).
        \item The instructions should contain the exact command and environment needed to run to reproduce the results. See the NeurIPS code and data submission guidelines (\url{https://neurips.cc/public/guides/CodeSubmissionPolicy}) for more details.
        \item The authors should provide instructions on data access and preparation, including how to access the raw data, preprocessed data, intermediate data, and generated data, etc.
        \item The authors should provide scripts to reproduce all experimental results for the new proposed method and baselines. If only a subset of experiments are reproducible, they should state which ones are omitted from the script and why.
        \item At submission time, to preserve anonymity, the authors should release anonymized versions (if applicable).
        \item Providing as much information as possible in supplemental material (appended to the paper) is recommended, but including URLs to data and code is permitted.
    \end{itemize}

\item {\bf Experimental setting/details}
    \item[] Question: Does the paper specify all the training and test details (e.g., data splits, hyperparameters, how they were chosen, type of optimizer) necessary to understand the results?
    \item[] Answer: \answerYes{}
    \item[] Justification: The experimental setup specifies the datasets, continual task ordering, backbone, optimizer, number of epochs, batch sizes, learning rate, LoRA rank, routing top-k, margin threshold, and evaluation metrics in Section \ref{sec:experiments} and Appendix \ref{app:implementation_details}.

    \item[] Guidelines:
    \begin{itemize}
        \item The answer \answerNA{} means that the paper does not include experiments.
        \item The experimental setting should be presented in the core of the paper to a level of detail that is necessary to appreciate the results and make sense of them.
        \item The full details can be provided either with the code, in appendix, or as supplemental material.
    \end{itemize}

\item {\bf Experiment statistical significance}
    \item[] Question: Does the paper report error bars suitably and correctly defined or other appropriate information about the statistical significance of the experiments?
    \item[] Answer: \answerNo{}
    \item[] Justification: The paper reports results across several datasets, backbones, and ablations, but does not currently include error bars, confidence intervals, or statistical significance tests for the main experiments.

    \item[] Guidelines:
    \begin{itemize}
        \item The answer \answerNA{} means that the paper does not include experiments.
        \item The authors should answer \answerYes{} if the results are accompanied by error bars, confidence intervals, or statistical significance tests, at least for the experiments that support the main claims of the paper.
        \item The factors of variability that the error bars are capturing should be clearly stated (for example, train/test split, initialization, random drawing of some parameter, or overall run with given experimental conditions).
        \item The method for calculating the error bars should be explained (closed form formula, call to a library function, bootstrap, etc.)
        \item The assumptions made should be given (e.g., Normally distributed errors).
        \item It should be clear whether the error bar is the standard deviation or the standard error of the mean.
        \item It is OK to report 1-sigma error bars, but one should state it. The authors should preferably report a 2-sigma error bar than state that they have a 96\% CI, if the hypothesis of Normality of errors is not verified.
        \item For asymmetric distributions, the authors should be careful not to show in tables or figures symmetric error bars that would yield results that are out of range (e.g., negative error rates).
        \item If error bars are reported in tables or plots, the authors should explain in the text how they were calculated and reference the corresponding figures or tables in the text.
    \end{itemize}

\item {\bf Experiments compute resources}
    \item[] Question: For each experiment, does the paper provide sufficient information on the computer resources (type of compute workers, memory, time of execution) needed to reproduce the experiments?
    \item[] Answer: \answerYes{}
    \item[] Justification: We discuss this in detail in \Cref{app:compute_resources}. We also specify the shared training configuration, including the CLIP ViT-B/16 backbone, number of epochs, batch size, optimizer, and other implementation details in \Cref{app:implementation_details}.
    \item[] Guidelines:
    \begin{itemize}
        \item The answer \answerNA{} means that the paper does not include experiments.
        \item The paper should indicate the type of compute workers CPU or GPU, internal cluster, or cloud provider, including relevant memory and storage.
        \item The paper should provide the amount of compute required for each of the individual experimental runs as well as estimate the total compute. 
        \item The paper should disclose whether the full research project required more compute than the experiments reported in the paper (e.g., preliminary or failed experiments that didn't make it into the paper). 
    \end{itemize}
    
\item {\bf Code of ethics}
    \item[] Question: Does the research conducted in the paper conform, in every respect, with the NeurIPS Code of Ethics \url{https://neurips.cc/public/EthicsGuidelines}?
    \item[] Answer: \answerYes{}
    \item[] Justification: The research uses existing public datasets and pretrained models for benchmark and method evaluation, and we are not aware of any aspect of the work that violates the NeurIPS Code of Ethics.

    \item[] Guidelines:
    \begin{itemize}
        \item The answer \answerNA{} means that the authors have not reviewed the NeurIPS Code of Ethics.
        \item If the authors answer \answerNo, they should explain the special circumstances that require a deviation from the Code of Ethics.
        \item The authors should make sure to preserve anonymity (e.g., if there is a special consideration due to laws or regulations in their jurisdiction).
    \end{itemize}

\item {\bf Broader impacts}
    \item[] Question: Does the paper discuss both potential positive societal impacts and negative societal impacts of the work performed?
    \item[] Answer: \answerYes{}
    \item[] Justification: The paper includes a broader impact discussion in \Cref{sec:conclusions}, where we discuss the responsible development of continual multimodal retrieval systems and acknowledge potential misuse risks associated with vision-language models and retrieval technologies.
    \Cref{sec:conclusions}.
    \item[] Guidelines:
    \begin{itemize}
        \item The answer \answerNA{} means that there is no societal impact of the work performed.
        \item If the authors answer \answerNA{} or \answerNo, they should explain why their work has no societal impact or why the paper does not address societal impact.
        \item Examples of negative societal impacts include potential malicious or unintended uses (e.g., disinformation, generating fake profiles, surveillance), fairness considerations (e.g., deployment of technologies that could make decisions that unfairly impact specific groups), privacy considerations, and security considerations.
        \item The conference expects that many papers will be foundational research and not tied to particular applications, let alone deployments. However, if there is a direct path to any negative applications, the authors should point it out. For example, it is legitimate to point out that an improvement in the quality of generative models could be used to generate Deepfakes for disinformation. On the other hand, it is not needed to point out that a generic algorithm for optimizing neural networks could enable people to train models that generate Deepfakes faster.
        \item The authors should consider possible harms that could arise when the technology is being used as intended and functioning correctly, harms that could arise when the technology is being used as intended but gives incorrect results, and harms following from (intentional or unintentional) misuse of the technology.
        \item If there are negative societal impacts, the authors could also discuss possible mitigation strategies (e.g., gated release of models, providing defenses in addition to attacks, mechanisms for monitoring misuse, mechanisms to monitor how a system learns from feedback over time, improving the efficiency and accessibility of ML).
    \end{itemize}
    
\item {\bf Safeguards}
    \item[] Question: Does the paper describe safeguards that have been put in place for responsible release of data or models that have a high risk for misuse (e.g., pre-trained language models, image generators, or scraped datasets)?
    \item[] Answer: \answerNA{}
    \item[] Justification: The paper does not release a high-risk generative model, scraped dataset, or other asset with a direct high risk for misuse; it releases an evaluation framework and method built on existing public assets.

    \item[] Guidelines:
    \begin{itemize}
        \item The answer \answerNA{} means that the paper poses no such risks.
        \item Released models that have a high risk for misuse or dual-use should be released with necessary safeguards to allow for controlled use of the model, for example by requiring that users adhere to usage guidelines or restrictions to access the model or implementing safety filters. 
        \item Datasets that have been scraped from the Internet could pose safety risks. The authors should describe how they avoided releasing unsafe images.
        \item We recognize that providing effective safeguards is challenging, and many papers do not require this, but we encourage authors to take this into account and make a best faith effort.
    \end{itemize}

\item {\bf Licenses for existing assets}
    \item[] Question: Are the creators or original owners of assets (e.g., code, data, models), used in the paper, properly credited and are the license and terms of use explicitly mentioned and properly respected?
    \item[] Answer: \answerYes{}
    \item[] Justification: The paper properly credits the pretrained models, datasets, and prior methods used throughout the experiments through citations and references, and uses publicly available assets under their respective terms of use.
    \item[] Guidelines:
    \begin{itemize}
        \item The answer \answerNA{} means that the paper does not use existing assets.
        \item The authors should cite the original paper that produced the code package or dataset.
        \item The authors should state which version of the asset is used and, if possible, include a URL.
        \item The name of the license (e.g., CC-BY 4.0) should be included for each asset.
        \item For scraped data from a particular source (e.g., website), the copyright and terms of service of that source should be provided.
        \item If assets are released, the license, copyright information, and terms of use in the package should be provided. For popular datasets, \url{paperswithcode.com/datasets} has curated licenses for some datasets. Their licensing guide can help determine the license of a dataset.
        \item For existing datasets that are re-packaged, both the original license and the license of the derived asset (if it has changed) should be provided.
        \item If this information is not available online, the authors are encouraged to reach out to the asset's creators.
    \end{itemize}

\item {\bf New assets}
    \item[] Question: Are new assets introduced in the paper well documented and is the documentation provided alongside the assets?
    \item[] Answer: \answerYes{}
    \item[] Justification: The paper releases a new evaluation framework and implementation through an anonymized repository. This repository serves as a practical documentation with setup instructions, dataset preparation, expected outputs, licenses, and limitations.

    \item[] Guidelines:
    \begin{itemize}
        \item The answer \answerNA{} means that the paper does not release new assets.
        \item Researchers should communicate the details of the dataset\slash code\slash model as part of their submissions via structured templates. This includes details about training, license, limitations, etc. 
        \item The paper should discuss whether and how consent was obtained from people whose asset is used.
        \item At submission time, remember to anonymize your assets (if applicable). You can either create an anonymized URL or include an anonymized zip file.
    \end{itemize}

\item {\bf Crowdsourcing and research with human subjects}
    \item[] Question: For crowdsourcing experiments and research with human subjects, does the paper include the full text of instructions given to participants and screenshots, if applicable, as well as details about compensation (if any)? 
    \item[] Answer: \answerNA{}
    \item[] Justification: The paper does not involve crowdsourcing or new research with human subjects.

    \item[] Guidelines:
    \begin{itemize}
        \item The answer \answerNA{} means that the paper does not involve crowdsourcing nor research with human subjects.
        \item Including this information in the supplemental material is fine, but if the main contribution of the paper involves human subjects, then as much detail as possible should be included in the main paper. 
        \item According to the NeurIPS Code of Ethics, workers involved in data collection, curation, or other labor should be paid at least the minimum wage in the country of the data collector. 
    \end{itemize}

\item {\bf Institutional review board (IRB) approvals or equivalent for research with human subjects}
    \item[] Question: Does the paper describe potential risks incurred by study participants, whether such risks were disclosed to the subjects, and whether Institutional Review Board (IRB) approvals (or an equivalent approval/review based on the requirements of your country or institution) were obtained?
    \item[] Answer: \answerNA{}
    \item[] Justification: The paper does not involve crowdsourcing or new research with human subjects, so IRB approval is not applicable.

    \item[] Guidelines:
    \begin{itemize}
        \item The answer \answerNA{} means that the paper does not involve crowdsourcing nor research with human subjects.
        \item Depending on the country in which research is conducted, IRB approval (or equivalent) may be required for any human subjects research. If you obtained IRB approval, you should clearly state this in the paper. 
        \item We recognize that the procedures for this may vary significantly between institutions and locations, and we expect authors to adhere to the NeurIPS Code of Ethics and the guidelines for their institution. 
        \item For initial submissions, do not include any information that would break anonymity (if applicable), such as the institution conducting the review.
    \end{itemize}

\item {\bf Declaration of LLM usage}
    \item[] Question: Does the paper describe the usage of LLMs if it is an important, original, or non-standard component of the core methods in this research? Note that if the LLM is used only for writing, editing, or formatting purposes and does \emph{not} impact the core methodology, scientific rigor, or originality of the research, declaration is not required.
    \item[] Answer: \answerNA{}
    \item[] Justification: The core method development and experiments do not involve LLMs as an important, original, or non-standard component.
    \item[] Guidelines:
    \begin{itemize}
        \item The answer \answerNA{} means that the core method development in this research does not involve LLMs as any important, original, or non-standard components.
        \item Please refer to our LLM policy in the NeurIPS handbook for what should or should not be described.
    \end{itemize}

\end{enumerate}

\end{document}